\documentclass[11pt,a4paper]{article}
\usepackage[hyperref]{acl2021}
\usepackage{times}
\usepackage{latexsym}

\usepackage{microtype}

\usepackage{amsmath}%
\usepackage{MnSymbol}%
\usepackage{wasysym}%
\usepackage{graphicx}
\usepackage{multirow}
\usepackage[linesnumbered,ruled,vlined,boxed,noend]{algorithm2e}
\SetAlFnt{\small}
\SetAlCapFnt{\normalsize}
\SetAlCapNameFnt{\normalsize}

\usepackage{etoolbox} % for algorithm2e comment
\makeatletter
\patchcmd{\@algocf@start}% <cmd>
  {-1.5em}% <search>
  {-0.3em}% <replace>
  {}{}% <success><failure>
\makeatother
\newcommand{\mydots}{\hbox to 1.1em{\hss.\hss.\hss.\hss}}
\newcommand{\cev}[1]{\reflectbox{\ensuremath{\vec{\reflectbox{\ensuremath{#1}}}}}}
\aclfinalcopy % Uncomment this line for the final submission
\title{Neural Combinatory Constituency Parsing}

\author{Zhousi Chen, Longtu Zhang, Aizhan Imankulova, and Mamoru Komachi\\
    Tokyo Metropolitan University \\
    6-6 Asahigaoka, Hino, Tokyo 191-0065, Japan \\
    \texttt{chen-zhousi@ed.tmu.ac.jp} \hspace{1em}
    \texttt{vincentzlt@outlook.com} \\
    \texttt{aizhan.imankulova@cogsmart-global.com} \hspace{1em}
    \texttt{komachi@tmu.ac.jp} \\
}

\date{}

\begin{document}
\maketitle
\begin{abstract}
    We propose two fast neural combinatory models for constituency parsing: binary and multi-branching.
    Our models decompose the bottom-up parsing process into 1) classification of
    tags, labels, and binary orientations or chunks and 2) vector composition based on the computed orientations or chunks.
    These models have theoretical sub-quadratic complexity and empirical linear complexity.
    The binary model achieves an F1 score of $92.54$ on Penn Treebank, speeding at $1327.2$ sents/sec.
    Both the models with XLNet provide near state-of-the-art accuracies for English.
    Syntactic branching tendency and headedness of a language are observed during the training and inference processes
    for Penn Treebank, Chinese Treebank, and Keyaki Treebank (Japanese).
    % $86.14$ , and $87.05$ on .
    % This model is approximately $6$ times faster than the model of \newcite{DBLP:conf/acl/ZhouZ19} under the same condition.
\end{abstract}

% \iftaclpubformat
\section{Introduction}

Transition-based and chart-based methods are two main paradigms for constituency parsing.
Transition-based parsers~\cite{DBLP:conf/naacl/DyerKBS16,DBLP:conf/acl/KitaevK20}
build a tree with a sequence of local actions.
Despite their $O(n)$ computational complexity,
the locality makes them less accurate and necessitates additional grammars or lookahead features for improvement~\cite{DBLP:conf/acl/KuhlmannGS11,DBLP:conf/acl/ZhuZCZZ13,DBLP:journals/tacl/LiuZ17b}.
By contrast, chart-based parsers are conceptually simple and accurate
when used with a CYK-style algorithm~\cite{DBLP:conf/acl/KleinK18,DBLP:conf/acl/ZhouZ19} for finding the global optima.
However, their complexity is $O(n^3)$.
To achieve both accuracy and simplicity (without high complexity) is a critical problem in parsing.

Recent efforts were made using neural models.
In contrast to earlier symbolic approaches \cite{DBLP:conf/anlp/Charniak00,DBLP:conf/acl/KleinM03}, 
neural models are simplified by utilizing their adaptive distributed representation,
thereby eliminating complicated symbolic engineering.
The seq2seq model for parsing~\cite{DBLP:conf/nips/VinyalsKKPSH15} leverages
such representation to interpret the structural task as a general sequential task.
With augmented data and ensemble, it outperforms the symbolic models mentioned in \newcite{DBLP:conf/acl/PetrovBTK06}
and provides a complexity of $O(n^2)$ with the attention mechanism~\cite{DBLP:journals/corr/BahdanauCB14}.
However, its performance is inferior to those of specialized neural parsers~\cite{DBLP:conf/iwpt/LiuZ17,DBLP:journals/tacl/LiuZ17a,DBLP:journals/tacl/LiuZ17b}.
\newcite{DBLP:conf/acl/SocherBMN13} proposed a parsing strategy for a symbolic constituent parser augmented with neural vector compositionality.
It did not outperform the two paradigms in neural style probably because the neural techniques, such as contextualization, are not fully exploited.
\newcite{DBLP:conf/acl/KitaevK20} showed that a simple transition-based model with
a dynamic distributed representation, BERT \cite{DBLP:conf/naacl/DevlinCLT19}, nearly delivers a state-of-the-art performance.

\begin{figure}[t]
    \centering
    \includegraphics[width=\linewidth]{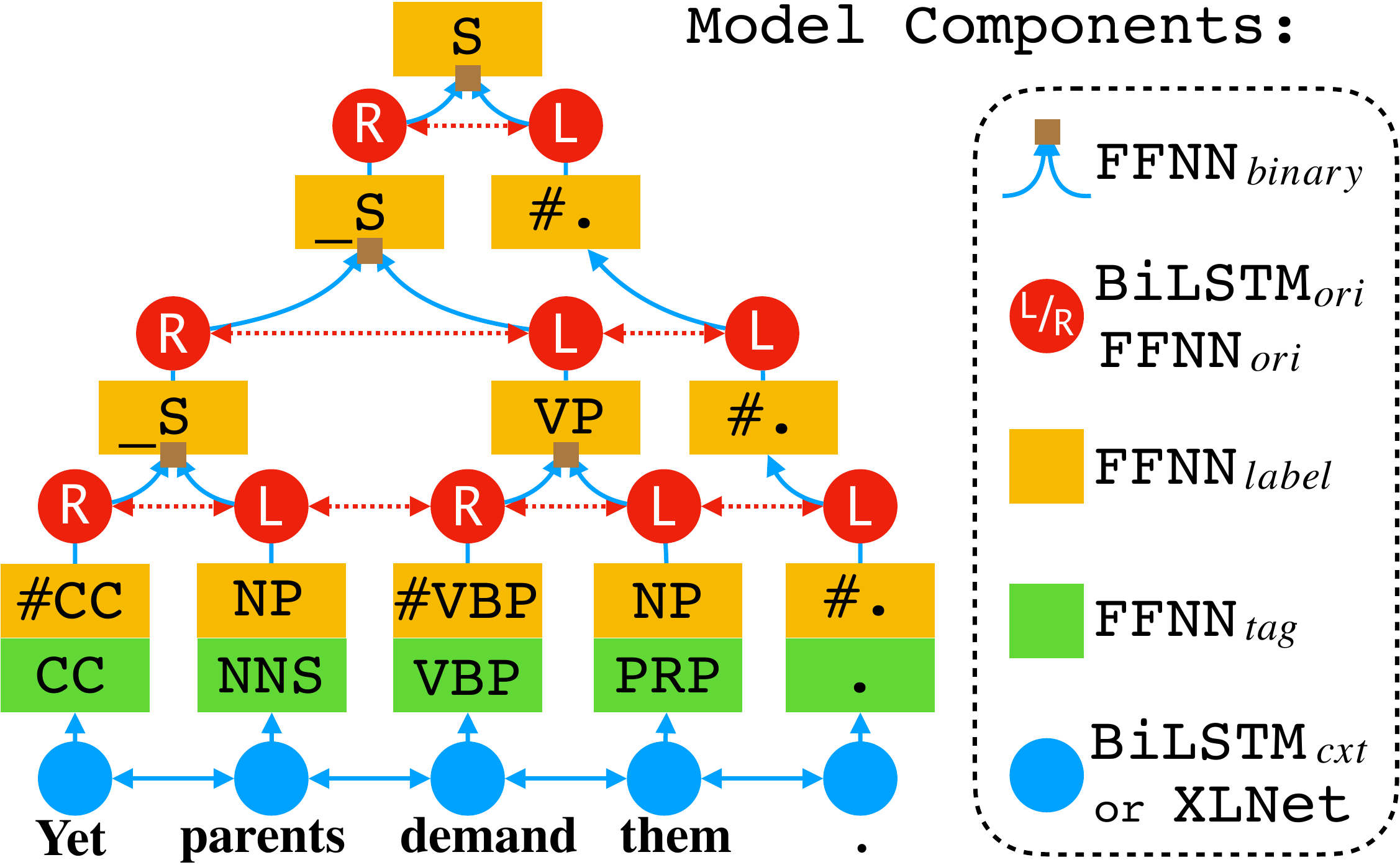}
    \caption{Parsing instance with the binary model.
    The bottom-up flow of word information
    is indicated by blue arrows and orientation flows by dotted red arrows. 
    Binary parsing explores the internal constituents of \texttt{S}. 
    Special labels prefixed with ``\#'' or ``\_'' are \textit{sub} category placeholders caused by binarization and stratification. \label{fig:model_bin}}
\end{figure}

\begin{figure}[t]
    \centering
    \includegraphics[width=\linewidth]{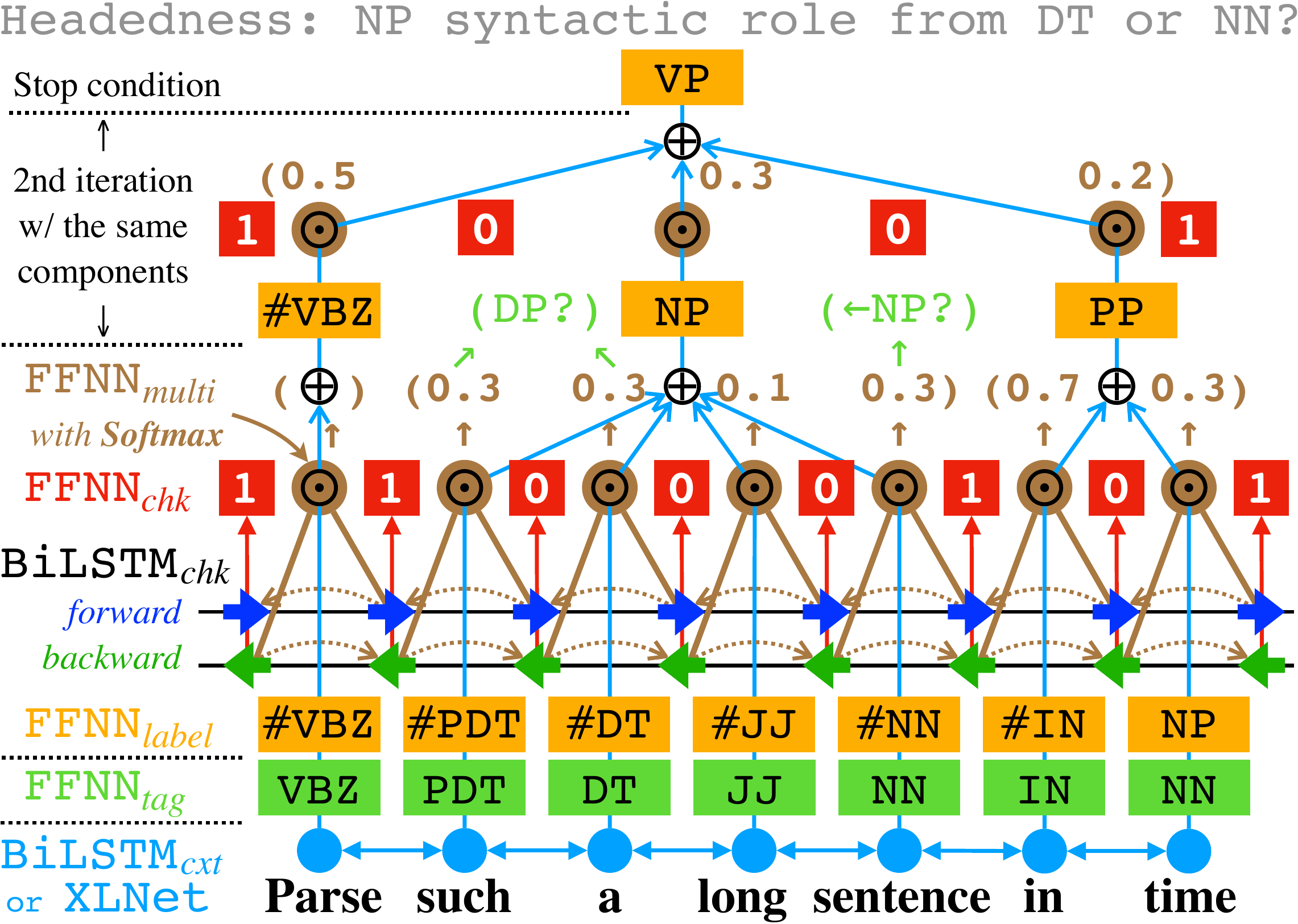}
    \caption{Multi-branching parsing uses chunks instead of orientations to form constituents.
    Chunks impose Softmax-normalized weights for their inputs.
    The unsupervised weights provide a shred of evidence for the headedness problem \cite{zwicky_1985}.
    \label{fig:model_mul}}
\end{figure}

We propose a pair of greedy combinatory parsers (i.e., neural combinators)
that efficiently utilize vector compositionality with recurrent components to address the aforementioned issues.
Their bottom-up parsing process is a recursive layer-wise loop
of classification and vector composition, as illustrated in Figures~\ref{fig:model_bin} \& \ref{fig:model_mul}.
Both parsers work on multiple unfolded variable-length layers,
iteratively combining vectors until one vector remains.
The binary model provides either left or right orientation for each word or constituent, whereas
the multi-branching model marks chunks as constituents at their boundaries.
Constituent embeddings are composed based on orientations or chunks.
Tagging and labeling are directly performed on all composed embeddings, creating the elements for building a tree: tags, labels, and paths.
The deterministic and greedy characteristics yield two simple and fast models, 
and they investigate different linguistic aspects.

The contributions of our study are as follows:
\begin{itemize}
    \item We propose two combinatory parsers\footnote{
        Our code, visualization tool, and pre-trained models are available at \url{https://github.com/tmu-nlp/nccp}
    } at $O(n)$ average-case complexity with a theoretical $O(n^2)$ upper bound.
    The binary parser achieves a competitive F1 score on Penn Treebank.
    Both models are the fastest and yet more compact than many previous models.
    \item We extend the proposed models with a recent pre-trained language model, XLNet \cite{DBLP:journals/corr/abs-1906-08237}.
    These models have higher speeds and are comparable to state-of-the-art parsers.
    \item The binary model leverages Chomsky normal form (CNF) factors as a training strategy and reflects the branching tendency of a language.
    The multi-branching model reveals constituent headedness \cite{zwicky_1985} with an attention mechanism.
\end{itemize}

\section{Previous Work}
\paragraph{Transition-based parsers.}
A transducer takes sequential lexical inputs and produces sequential tree-constructing actions in $O(n)$ time.
Although it can perfectly parse formal languages, complex semantics and long dependencies make it difficult to parse natural languages.
Informative features \cite{DBLP:journals/tacl/LiuZ17b,DBLP:conf/acl/KitaevK20,DBLP:conf/nips/YangD20}, 
or training and decoding strategies such as dynamic oracles \cite{DBLP:conf/emnlp/CrossH16},
reranking \cite{DBLP:conf/acl/CharniakJ05}, beam search, and ensemble,
can increase the accuracy.
However, these make the models complex, and the paradigm fails to naturally parallelize actions.
% Moreover, the sequential style prevents the parallelism of actions from accelerating a parse.

\paragraph{Chart-based parsers.} An exhaustive search algorithm checks every possibility
in a triangular chart and finds the optimal tree globally.
Recent neural chart parsers have achieved state-of-the-art accuracy
\cite{DBLP:conf/acl/KleinK18,DBLP:conf/acl/ZhouZ19,DBLP:conf/emnlp/MriniDTBCN20,DBLP:conf/ijcai/ZhangZL20}.
Despite their high accuracy, they are comparatively inefficient. 
Only $2n-1$ of $O(n^2)$ scoring nodes in the chart contain true constituents; many are filler nodes.
Chart parsers are often specially engineered for high-speed decoding. (e.g., using Cython)

\paragraph{Other parsers.} \citet{DBLP:conf/acl/BengioSCJLS18} and \citet{DBLP:conf/acl/NguyenNJL20} proposed 
local-and-greedy parsers in the top-down splitting style.
Their models facilitate divide-and-conquer algorithms that construct the tree based on the magnitude of the splitting scores.
A similar way of leveraging concurrent and greedy operations appears in an easy-first parser \cite{DBLP:conf/naacl/GoldbergE10}.
Sequential labeling \cite{DBLP:conf/emnlp/Gomez-Rodriguez18,DBLP:conf/acl/WeiWL20} is a new active thread that
also enables parallelism and fast decoding.
\citet{DBLP:journals/jmlr/Collobert11} designed an iterative chunking process for parsing.
His work stratifies trees into levels of IOBES prefixed constituent chunking nodes.
% The  in a Graph Transformer Network.
Similar to ours, his parser works from the bottom levels to higher levels.
However, the complexity is fixed at $O(n^2)$ without any node combinations.
All models introduced in this section do not exploit vector compositionality.

\section{Neural Combinatory Parsing}
\subsection{Data and Complexity} \label{sec:data}
Our models require stratified trees to train recurrent layers, 
and the binary model requires further binarization.
Stratification and binarization introduce redundant relaying nodes to the trees.

\begin{figure}[t]
    \centering
    \includegraphics[width=\linewidth]{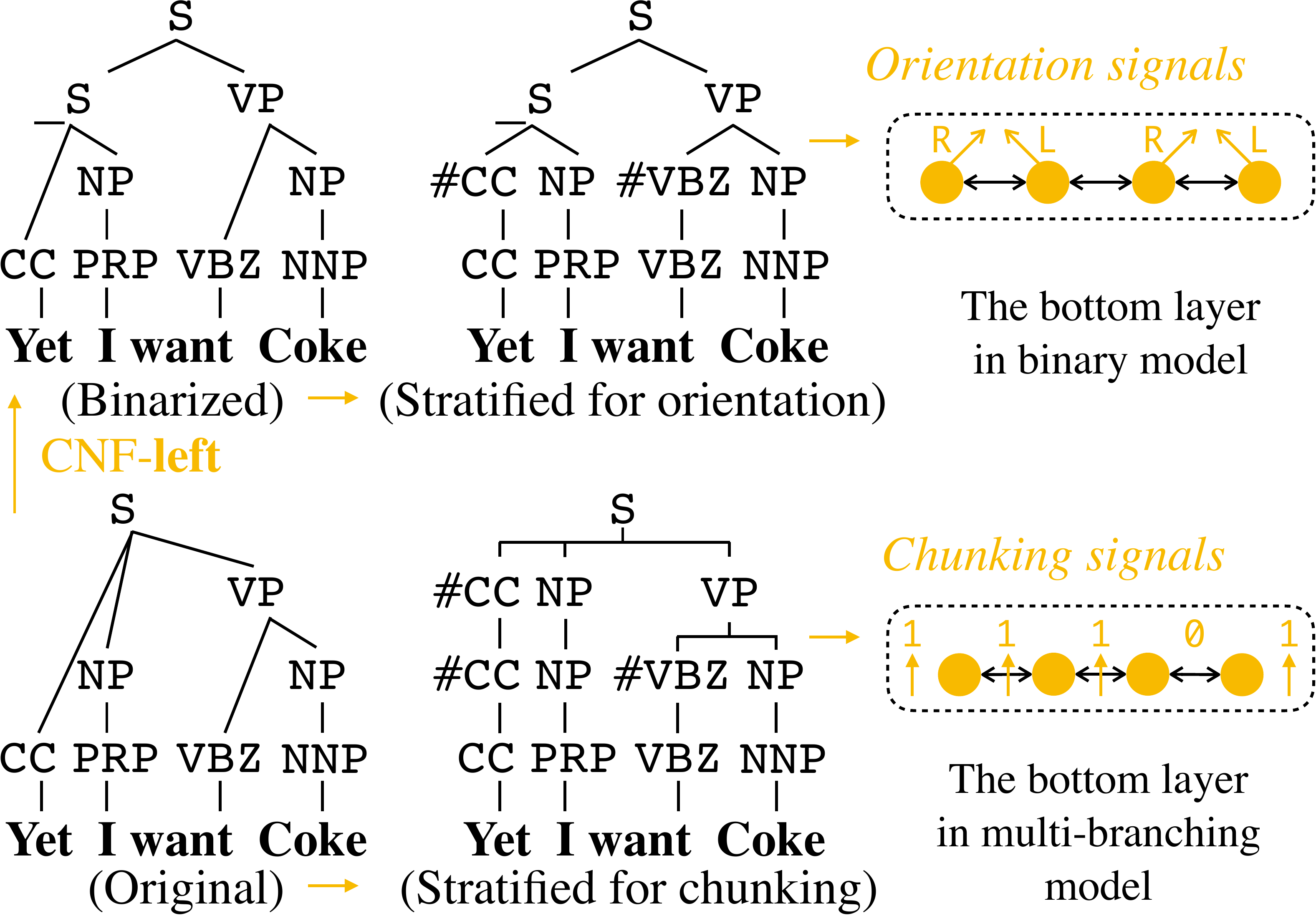}
    \caption{Example illustrating our tree stratification. 
    Both binary and multi-branching stratified trees contain redundancy,
    from which the original tree can be recovered with a few simple heuristic rules.
    Four in a layer combine into two with a compression ratio of 1/2.\label{fig:data}}
\end{figure}

\paragraph{Tree binarization.}
From the bottom-up perspective, a binary tree describes the order 
in which words and constituents combine with their neighbors into larger constituents, as shown in Figure \ref{fig:data}.
The orientations of the four words (i.e., right-left-right-left) determine the first combination.

\begin{table}[t]
    \begin{center}
        \scalebox{.9}{
        \begin{tabular}{c c c}
        % \toprule
        \hline
        \textbf{Category} & \textbf{Samples} & \textbf{\# Types} \\
        \hline
        % \midrule
        
        \textbf{Original} & \texttt{S} \texttt{NP} \texttt{VP} \texttt{SBAR+S} & $104$ \\
        \textbf{\_Sub} & \texttt{\_S} \texttt{\_NP} \texttt{\_VP} \texttt{\_PP} & $25$ \\
        \textbf{\#POS} & \texttt{\#NNP} \texttt{\#DT} \texttt{\#JJ} \texttt{\#.} & $45$ \\
        % \bottomrule
        \hline
        \end{tabular}
        }
    
        \caption{Three categories of our constituent label set with their samples and number of types.
        This is created from Penn Treebank (PTB). `\textbf{\_Sub}' and `\textbf{\#POS}' are relaying types, 
        which we group into a \textit{sub} category.\label{tab:label}}
    \end{center}
\end{table}

After binarization, we label the relaying sub-constituents with the parent label prefixed with an underscore mark.
If terminal POS tags do not immediately form constituents,
we create relaying placeholders prefixed with a hash mark\footnote{
    Multi-branching trees do not require binarization.
    The `\textbf{\_Sub}' group disappears, but the `\textbf{\#POS}' group persists.
}, as presented in Table \ref{tab:label}.
Unary branches were collapsed into a single node. Plus marks were used to join their labels (e.g., \texttt{SBAR+S}),
and all trace branches were removed.
The CNF with either a \textbf{left} or a \textbf{right} factor is commonly used.
However, it is heuristically biased, and trees can be binarized using other balanced splits
such as always splitting from the center to create a complete binary tree (\textbf{mid-out}) and
iteratively performing \textbf{left} and \textbf{right} to create another balanced tree (\textbf{mid-in}).
Finally, the orientation is extracted from the paths of these binary trees.

\begin{table}[t]
    \begin{center}
    \scalebox{.9}{
    \begin{tabular}{c c c c c}
    \hline
    \textbf{CNF} & \multicolumn{2}{c}{\textbf{Left-factoring}} & \multicolumn{2}{c}{\textbf{Right-factoring}} \\
    \textbf{Ori.} & \textbf{Left} & \textbf{Right} & \textbf{Left} & \textbf{Right} \\
    \hline
    \textbf{PTB} & 3.8M & \underline{4.4M} & 2.3M & \underline{6.5M} \\
    \textbf{CTB} & \underline{2.5M} & 1.7M & 1.4M & \underline{2.8M} \\
    \textbf{KTB} & \underline{4.5M} & 0.9M & 1.8M & \underline{2.1M} \\
    \hline
    \textbf{nCNF} & \multicolumn{2}{c}{\textbf{Midin-factoring}} & \multicolumn{2}{c}{\textbf{Midout-factoring}} \\
    \hline
    \textbf{PTB} & 3.0M & \underline{5.3M} & 2.8M & \underline{5.2M} \\
    \textbf{CTB} & 1.9M & \underline{2.2M} & 1.7M & \underline{2.1M} \\
    \textbf{KTB} & \underline{2.8M} & 1.7M & \underline{2.5M} & 1.2M \\
    \hline
    \end{tabular}
    }
    \end{center}
    \caption{Frequencies of orientation with different CNF (biased)
    and non-CNF (balanced) factors in different stratified corpora. \label{tab:ori}}
\end{table}

\begin{figure}[t]
    \centering
    \includegraphics[width=\linewidth]{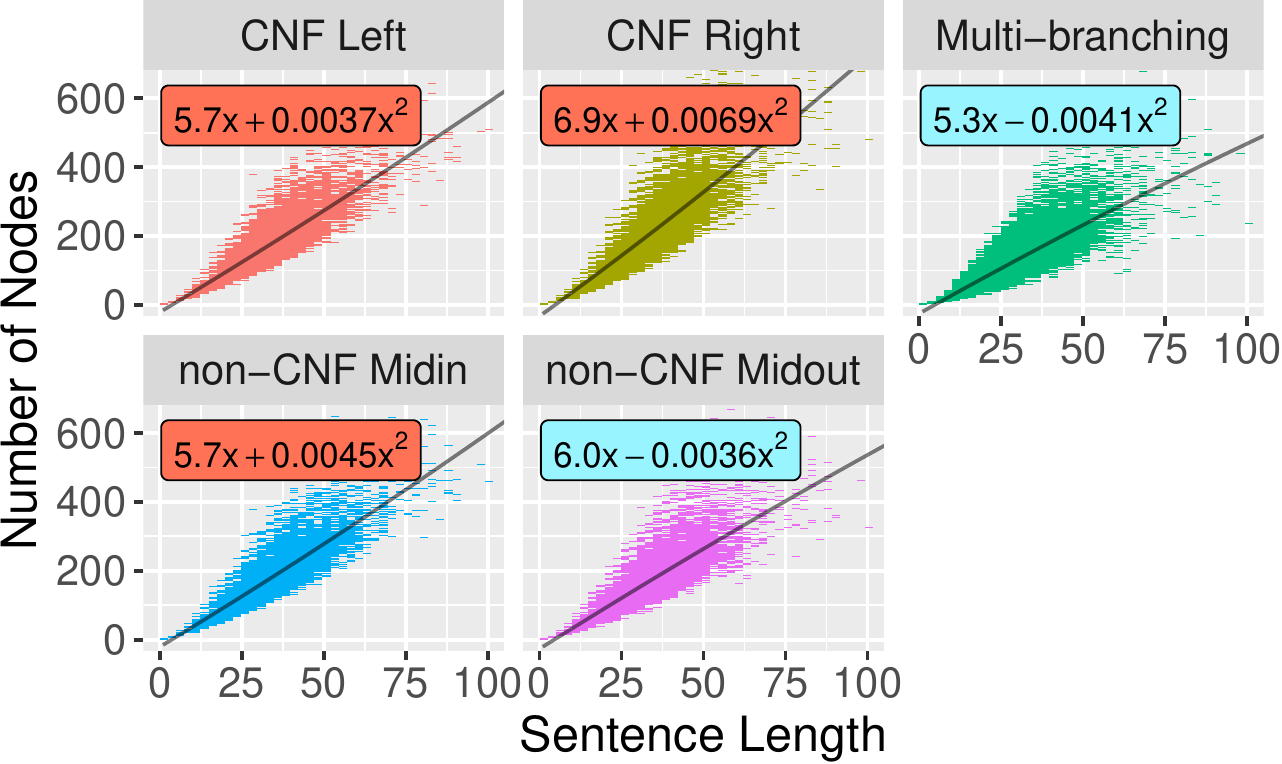}
    \caption{Linear empirical complexity in stratified PTB. 
    Linear regression reflects insignificant $O(n^2)$ tendencies.
    We differentiated the quadratic terms with red or light blue colors and omitted the constant biases. \label{fig:complexity}}
\end{figure}

We binarized Penn Treebank \cite[PTB]{DBLP:journals/coling/MarcusSM94} for English, 
Chinese Treebank \cite[CTB]{DBLP:journals/nle/XueXCP05} for Chinese, 
and Keyaki Treebank\footnote{\url{https://github.com/ajb129/KeyakiTreebank/tree/master/treebank}} \cite[KTB]{KTB} for Japanese to present
the syntactic branching tendencies in Table \ref{tab:ori}.
As English is a right-branching language, its majority orientation is to the right.
Even left-factoring cannot reverse the trend, but it should create a greater balance.
Figure \ref{fig:complexity} shows that it is less effective to stratify PTB with a \textbf{right} factor because it enhances the tendency.
The reverse tendency emerges in the KTB corpus as Japanese is a left-branching language.
For Chinese, CTB does not exhibit a clear branching tendency.
Non-CNF factors preserve the original tendency.

\begin{figure}[t]
    \centering
    \includegraphics[width=\linewidth]{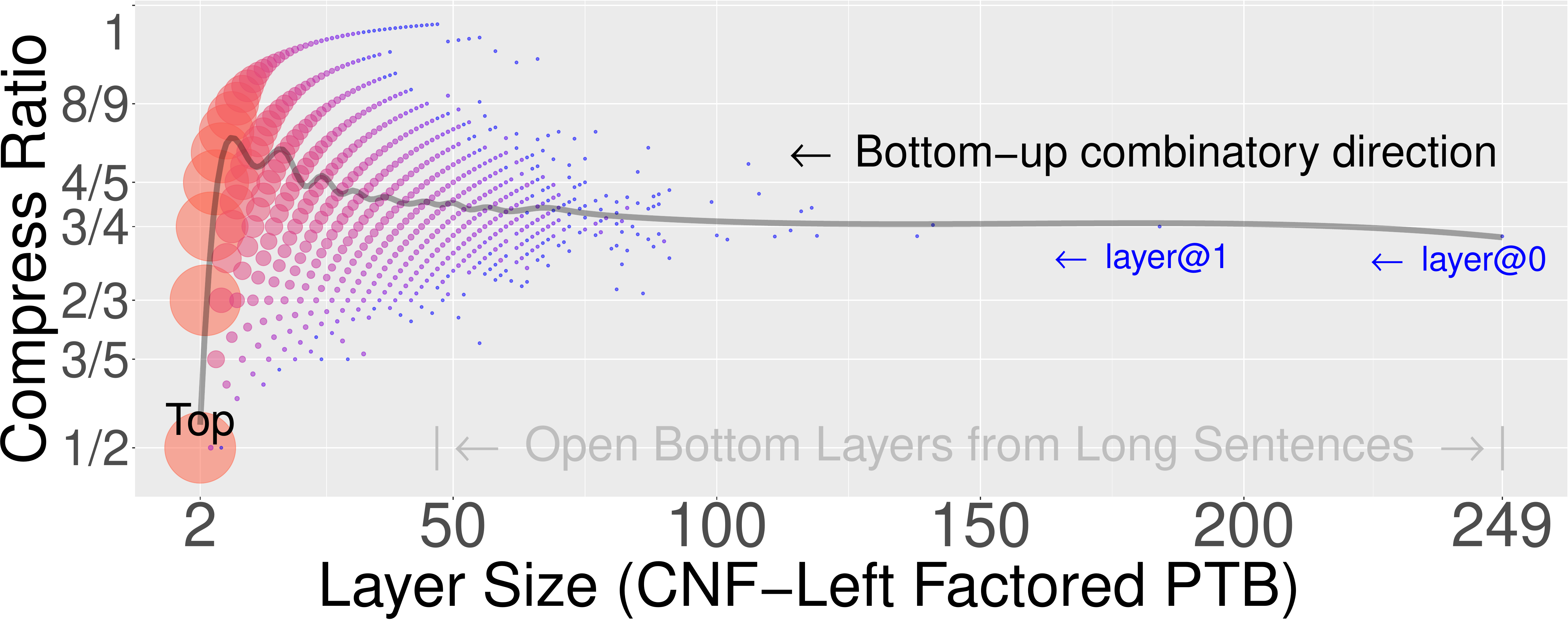}
    \caption{Layer-wise compression ratio over different sizes.
    The dot size was proportional to the situation count.
    % The ratio rarely fell to 0.5 unless it was the top layer.
    Statistically, the ratio had a stable mean of 0.77. \label{fig:level_ratio}}
\end{figure}

\paragraph{Complexity.}
Our models are trained with stratified treebanks.
The complexity for inference follows the total number of nodes in each layer of a tree.
There are two ideal cases: 1) Complete balanced trees with complexity $O(n)$.
They contain multiple independent phrases and enable full concurrency.
2) Trees with a single dependency core.
The model reduces a constant number of nodes in each layer, resulting in $O(n^2)$ complexity.

While each parse is a mixture of many cases, the \textit{empirical} complexity prefers the first case.
Formally, the average-case complexity can be inferred as $O(n)$ with the help of a stable compression ratio $0 < C <1$
($C\ge 0.5$ for binary).
Let $m_i$ represent the number children of the $i$-th tree in a general layer;
the compression ratio can be stated as $C = \frac{\sum_i 1}{\sum_i m_i}$.
Our stratified treebanks give stable $C$s for layers of different lengths, as shown in Figure \ref{fig:level_ratio}.
For the $k$-th layer of a sentence with $n$ words,
the number of nodes to compute can be expected to be $C^k \cdot n$.
Based on tree height $K > 0$, the expected number of total parsing nodes is
\begin{equation*}
    \sum^{K}_{k=0} C^k \cdot n
    = n \cdot \sum_{k=0}^{\infty} C^k - n \cdot \sum_{k=K+1}^{\infty} C^k
    < \frac{n}{1-C} \rlap{.}
\end{equation*}
% Because of the small and fixed influence of top layers,
The partial geometric series determines an empirically linear complexity on average.
% It is necessary that the multi-branching treebanks have fewer nodes.
% It also implies the reason that treebanks have a stable compression ratio: the tree structure is one-to-many.

\textit{Theoretically}, the complexity has a quadratic upper bound.
The general layer with
\begin{equation*}
    m_i =
    \begin{cases}
        M & \text{if } i = j \\
        1 & \text{otherwise}
    % \rlap{,}
    \end{cases}
\end{equation*} 
entails the second case, where $m_j$ is the only $M$-ary branch in each layer.
The nodes shape a triangular stratified tree with an $O(n^2)$ complexity.
However, this case is rare, especially for long sentences that should contain several concurrent phrases.
Otherwise, regression in Figure \ref{fig:complexity} should show significant $O(n^2)$ tendencies.
(See Appendix \ref{sec:lang_ratio} for more support and examples.)

\paragraph{Data structure.}
To summarize the data components of a treebank corpus, we used four tensors of indices for
1) words,
2) POS tags,
3) stratified syntactic labels,
4) stratified orientations, or 
5) stratified chunks,
$$\left(x_{0:n}, t_{0:n}, l^{0:k}_{0:n_k}, o^{0:k}_{0:n_k}\ \mathit{or}\ \, c^{0:k}_{0:n_k+1}\right)_j \in D,$$
where $n$ is the length of the $j$-th sentence,
$k$ indicates the $k$-th layer of the stratified data,
and $n_k$ is the layer length.
``:'' indicates a range of a sequence.

\subsection{Combinatory Parsing} \label{sec:model}

\begin{algorithm}[t]
    \SetKwFunction{BiLSTMcxt}{BiLSTM$_{cxt}$}
    \SetKwFunction{BiLSTMori}{BiLSTM$_{ori}$}
    \SetKwFunction{CrossEntropy}{CROSS-ENTROPY}
    \SetKwFunction{FFNN}{FFNN}
    \SetKwFunction{Compose}{COMPOSE}
    \SetKwFunction{Main}{PARSE}
    \SetAlgoLined
    \DontPrintSemicolon
    % \SetNoFillComment
    % \LinesNotNumbered
    \SetInd{0.0em}{1.2em}
    \SetKwProg{Mn}{Function}{:}{}
    \Mn{\Main{$e_{0:n}; \ t_{0:n},\ l^{0:k}_{0:n_k},\ o^{0:k}_{0:n_k}$\ or\ \, $c^{0:k}_{0:n_k+1}$}}{
        $x_{0:n}^0 \gets$ \BiLSTMcxt{$e_{0:n}$}\;
        \For{$i\gets0$ \KwTo $n-1$}{
            $\hat{t}_i \gets$ \FFNN$_{tag}${($x_i^0$)}\;
            $L_{tag} \gets$ \CrossEntropy{$t_i, \hat{t_i}$}\;
            }
            
        \For{$j\gets0$ \KwTo $k$}{
            \For{$i\gets0$ \KwTo $n_j-1$}{
                $\hat{l}_i^j \;\gets$ \FFNN$_{label}${($x_i^j$)}\;
                $L_{label} \gets$ \CrossEntropy{$l_i^j,\hat{l}_i^j$}\;
            }
            $x_{0:n_{j+1}}^{j+1} \gets$ \Compose{$x^{j}_{0:n_j};\ o^{j}_{0:n_j}\ \mathit{or}\ \, c^{j}_{0:n_j+1}$}\;
        }
        \KwRet $\hat{t}_{0:n}, \hat{l}^{0:k}_{0:n_k}$\; %, x^{0:k}_{0:n_k}, L_{tag}, L_{label}
    }

    \caption{Combinatory Parsing\label{alg:enc}}
\end{algorithm}

Our models comprise four feedforward (FFNN) 
and two bidirectional LSTM (BiLSTM) networks to decompose parsing into collaborative functions,
as shown in Algorithm \ref{alg:enc}.
During training, we use teacher forcing.
In the inference phase, the supervised signals behind all semicolons are ignored; the predicted signals serve as their substitute.

Input $e_{0:n}$ is an embedding sequence indexed by $x_{0:n}$.
In lines 2--5, the model prepares a contextual sequence for the combinator and predicts the lexical tags.
Lines 6--10 describe the layer-wise loop of the combinator.

The tagging and labeling functions, \texttt{FFNN}$_{tag}$ and \texttt{FFNN}$_{label}$, are 2-layer FFNNs.
Their first layer is shared,
creating a hidden layer necessary for projecting diversified situations in the manifold to the non-zero logits for the argmax decision.
The core function \texttt{COMPOSE}\footnote{\texttt{COMPOSE} with BiLSTM cannot be parallelized to $O(1)$.} is either a binary Algorithm \ref{alg:enc_bin} or a multi-branching Algorithm \ref{alg:m_ary}.

% The total number of the sequential  operations in  (i.e., layers) is the complexity.

\begin{algorithm}[t]
    \SetKwFunction{BiLSTMori}{BiLSTM$_{ori}$}
    \SetKwFunction{FFNNori}{FFNN$_{ori}$}
    \SetKwFunction{FFNNitp}{FFNN$_{binary}$}
    \SetKwFunction{HingeLoss}{HINGE-LOSS}
    \SetKwFunction{Compose}{COMPOSE}
    \SetKwFunction{Binary}{BINARY}
    \SetAlgoLined
    \DontPrintSemicolon
    % \SetNoFillComment
    % \LinesNotNumbered
    \SetInd{0.0em}{1.2em}
    \SetKwProg{Mn}{Function}{:}{}
    \SetKwProg{Fn}{Function}{:}{}
    \Mn{\Compose{$x_{0:n_j}^{j}; \ o_{0:n_j}^{j}$}}{ % $o_L, o_R,\ x_L, x_R; \text{Var}$
        $h_{0:n_j}^{j} \gets$ \BiLSTMori{$x_{0:n_j}^{j}$}\;
        \For{$i\gets0$ \KwTo $n_j-1$}{ % (\tcp*[f]{$o^{j}_{0:n_j}$})
            $\hat{o}_i^j \:\gets$ \FFNNori{$h_i^j$}\;
            $L_{ori} \gets$ \HingeLoss{$o_i^j, \hat{o}_i^j$}\;
            \uIf{$i > 0$ and $\hat{o}_{i-1}^j + (1 - \hat{o}_i^j) > 0$}{
                Append \Binary{$\hat{o}_{i-1}^j, \hat{o}_i^j,\ x_{i-1}^j, x_i^j$} to $x^{j+1}$\;
            }
        }
        \KwRet $x_{0:n_{j+1}}^{j+1}$\; % ; \ o_{0:n_{j+1}}^{j+1}, L_{ori}
        }
    \;
    \SetKwProg{Fn}{Function}{:}{}
    \Fn{\Binary{$o_L, o_R,\ x_L, x_R$}}{
        \uIf(\tcp*[f]{relay}){$o_L + (1-o_R) = 1$}{
            \KwRet $o_L \cdot x_L + (1-o_R) \cdot x_R$\;
        }\uElse(\tcp*[f]{vector interpolation}){
            $\lambda \gets \sigma$ \FFNNitp{$x_L \oplus x_R$}\;
            \KwRet $\lambda \odot x_L + (1-\lambda) \odot x_R$\;
        }
    }
\caption{Binary Compose\label{alg:enc_bin}}
\end{algorithm}

\paragraph{Binary model.}
In Algorithm \ref{alg:enc_bin}, the orientation function is hinted by \texttt{BiLSTM}$_{ori}$.
A single-layer \texttt{FFNN}$_{ori}$ with a threshold
reduces the outputs to an integer of either 0 or 1 to indicate two possible orientations.
In function \texttt{BINARY}, when two adjacent orientations agree as they sum to 2, 
their embeddings are combined by a combinatory operation.
$\sigma$ is the Sigmoid function, ``$\oplus$'' represents concatenation, and ``$\odot$'' represents pointwise multiplication.
(See Appendix \ref{sec:bi} for more binary variants.)

\begin{algorithm}[t]
    \SetKwFunction{BiLSTMchk}{BiLSTM$_{chk}$}
    \SetKwFunction{FFNNchk}{FFNN$_{chk}$}
    \SetKwFunction{FFNNitp}{FFNN$_{multi}$}
    \SetKwFunction{Compose}{COMPOSE}
    \SetKwFunction{Multi}{MULTI}
    \SetAlgoLined
    \DontPrintSemicolon
    \SetInd{0.0em}{1.2em}
    \SetKwProg{Mn}{Function}{:}{}
    \SetKwProg{Fn}{Function}{:}{}
    \Mn{\Compose{$x_{0:n_j}^{j}; \ c_{0:n_j+1}^{j}$}}{
        $\vec{h}_{0:n_j}^{j}, \cev{h}_{0:n_j}^{j} \gets$ \BiLSTMchk{$x_{0:n_j}^{j}$}\;
        Pad $\vec{h}_{0:n_j}^{j}$ with $\vec{h}^{j}_{-1}$ and $\cev{h}_{0:n_j}^{j}$ with $\cev{h}^{j}_{n_j}$\;
        \For{$i\gets0$ \KwTo $n_j$}{ % (\tcp*[f]{$c^{j}_{0:n_j+1}$})
            $\hat{c}_i^j \:\gets$ \FFNN$_{chk}${($\vec{h}_{i-1}^j \oplus \cev{h}_i^j$)}\;
            $L_{chk} \gets$ \HingeLoss{$c_i^j, \hat{c}_i^j$}\;
            \uIf(Append $i$ to $s$\tcp*[f]{segment}){$\hat{c}_i^j=1$}{}
            \uIf{$i<n_j$}{ % (\tcp*[f]{$c^{j}_{0:n_j+1}\,$longer than$\ d^{j}_{0:n_j}$})
                $d^j_i \gets$ $(\vec{h}^{j}_{i} - \vec{h}^{j}_{i-1}) \oplus (\cev{h}^{j}_{i} - \cev{h}^{j}_{i+1})$\;
            }
        }
        \For{$i\gets 0$ \KwTo $ \vert s \vert - 1$}{ % (\tcp*[f]{N-ary COMPOSE})
            Append \Multi{$d^j_{s_i:s_{i+1}},\ x^j_{s_i:s_{i+1}}$} to $x^{j+1}$\;
        }
        \KwRet $x_{0:n_{j+1}}^{j+1}$\; % ; \ c_{0:n_{j+1}}^{j+1}, L_{chk}
    }
    \;
    \SetKwProg{Fn}{Function}{:}{}
    \Fn{\Multi{$d_{chk},\ x_{chk}$}}{
        $\lambda_{chk} \gets \mathit{Softmax}($ \FFNNitp{$d_{chk}$}$)$\;
        \KwRet $\sum_i^{chk} \lambda_i \odot x_{i}$
    }
    \caption{Multi-branching Compose\label{alg:m_ary}}
\end{algorithm}

\paragraph{Multi-branching model.} To resemble binary interpolation,
we use the Softmax function for each chunk, as described in Algorithm \ref{alg:m_ary}.
\texttt{BiLSTM}$_{chk}$ is in place of \texttt{BiLSTM}$_{ori}$ to hint \texttt{FFNN}$_{chk}$ emitting chunk signals.
% $d^j_{0:n_j}$ is the concatenation of the forward and backward differences of $h^j_{0:n_j}$ padded with $\vec{h}^{j}_{-1}$ and $\cev{h}^{j}_{n_j}$.
Segment $s$ splits $x^j_{0:n_j}$ and $d^j_{0:n_j}$ into chunks of $x_{chk}$ and $d_{chk}$.
\texttt{FFNN}$_{multi}$ and Softmax turn $d_{chk}$ into attention $\lambda_{chk}$ to interpolate vector chunk $x_{chk}$.
Binary interpolation $\lambda$ is a special case of the multi-branching $\lambda_{chk}$
because Sigmoid and Softmax functions are closely related.

To obtain the final tree representation, we apply a symbolic pruner in the same bottom-up manner
to remove redundant nodes, expand the collapsed nodes, and assemble the sub-trees based on the neural outputs.
(See Appendix \ref{sec:symbolic}.)

\section{Experiments}

We follow previous data splits for PTB, CTB, and KTB (See Appendix \ref{sec:exp_set}).
The preprocessing of data is described in Section~\ref{sec:data}.

For the binary model,
we explored interpolated dynamic datasets by sampling two CNF factored datasets.
This is because of the following: 1) The experiments with the non-CNF factors did not yield any promising results; thus, we have not reported them.
2) The language was loosely left-branched, right-branched, or did not show a noticeable tendency.
Moreover, the use of a single static dataset may introduce a severe orientation bias. %, which may be harmful to the model;
3) All factors are intermediate variables and equally correct.
We defined the sampling strategies with two static CNF-factored datasets at certain ratios
and named each strategy in the format ``L\%R\%'' according to the ratio percentages.
Our experiments mainly focus on binary model \textbf{B}
because of the aforementioned property for training parsers more accurate than multi-branching model \textbf{M}.

Our parsers do not contain lexical information components \cite{DBLP:journals/tacl/LiuZ17b,DBLP:conf/acl/KleinK18}.
Instead, we use fastText \cite{DBLP:journals/tacl/BojanowskiGJM17}
because we can obtain pre-trained models easily for many languages or train new ones from scratch with the corpora at hand.
We examined its influence in Section \ref{sec:abl}, whereas the official pre-trained embeddings are the default.

Meanwhile, pre-trained language models are useful for various tasks, 
including constituency parsing~\cite{DBLP:conf/acl/KleinK18,DBLP:conf/acl/KitaevK20,DBLP:conf/acl/ZhouZ19,DBLP:conf/nips/YangD20,DBLP:conf/emnlp/MriniDTBCN20}.
We chose XLNet~\cite{DBLP:journals/corr/abs-1906-08237} to compare with the static fastText embeddings.
Specifically, either a 1-layer FFNN (/0) or an $n$-layer BiLSTM (/$n^+$) was used to convert the 768-unit output to our model size.
We used a GeForce GTX 1080 Ti with 11 GB and a TITAN RTX with 24GB memory only for tuning XLNet.

The model size for vector compositionality was set at 300.
The hidden sizes for labeling, orientation, and chunking were 200, 64, and 200, respectively.
Different numbers of layers of the \texttt{BiLSTM}$_{cxt}$ (/$n$) were explored, and the default was six layers.
% To further investigate the orientation function, we compared a bidirectional quasi-recurrent neural network
% \cite[BiQRNN]{DBLP:conf/iclr/0002MXS17} with \texttt{BiLSTM}$_{ori}$.
% This network is a member of the RNN family with a simpler gating mechanism.
% We further simplified it by tuning off gating and relaying its state at \texttt{<nil>} nodes to investigate their effect on the model.
\texttt{HINGE-LOSS} was the default criterion for orientation while binary cross-entropy (\texttt{BCE-LOSS}) was tested.
The coefficients of the three losses were explored and the default were $L = 0.2 \cdot L_{tag} + 0.3 \cdot L_{label} + 0.5 \cdot L_{ori\ (\mathbf{or}\ chk)}$.

\subsection{Overall Results} \label{sec:ovr}
\begin{table*}[t]
    \begin{center}
        \setlength\tabcolsep{5pt}
        \scalebox{.9}{
        %\begin{tabular}{p{2.43cm} l c c c}
        \begin{tabular}{l | l c c c c | l c c c}
        \hline
        \textbf{Corpus} & \multicolumn{5}{c |}{\textbf{Penn Treebank}} & \multicolumn{4}{c}{\textbf{Chinese Treebank}} \\
        \textbf{Single Model} & \textbf{Type} & \textbf{sents/sec} & \textbf{LP} & \textbf{LR} & \textbf{F1} & \textbf{Type} & \textbf{LP} & \textbf{LR} & \textbf{F1}\\
        \hline
        % \shortcite{DBLP:conf/nips/VinyalsKKPSH15} & S2S & - & - & 88.3 \\
        % \newcite{DBLP:conf/naacl/DyerKBS16} & T$\downarrow$ & - & - & - & 89.8 & T$\downarrow$ & - & - & 84.6\\
        % \newcite{DBLP:conf/acl/ZhuZCZZ13} & T$\uparrow$ (16) & 89.5 & 90.7 & 90.2 & 90.4 & T$\uparrow$ (16) & 84.3 & 82.1 & 83.2 \\
        % \newcite{DBLP:conf/acl/SocherBMN13} & O & - & - & - & 90.4 & - & - & - & - \\
        \newcite{DBLP:conf/acl/WatanabeS15} & T$\uparrow$ (32) & - & - & - & 90.7 & T$\uparrow$ (64) & - & - & 84.3 \\
        \newcite{DBLP:conf/emnlp/Gomez-Rodriguez18} & O & \underline{898} & - & - & 90.7 & O & - & - & 83.1 \\
        \newcite{DBLP:conf/emnlp/CrossH16} & T$\uparrow$ (1) & - & 92.1 & 90.5 & 91.3 & - & - & - & - \\
        \newcite{DBLP:journals/tacl/LiuZ17b} & T$\downarrow$ (16) & 79.2 & 92.1 & 91.3 & 91.7 & T$\downarrow$ (16) & 85.9 & 85.2 & 85.5 \\
        \newcite{DBLP:conf/acl/SternAK17} & C & 75.5 & 93.0 & 90.6 & 91.8  & - & - & - & - \\
        \newcite{DBLP:conf/acl/BengioSCJLS18} & O$\downarrow$ (1) & 111.1 & 92.0 & 91.7 & 91.8 & O$\downarrow$ (1) & 86.6 & 86.4 & 86.5 \\
        \newcite{DBLP:conf/acl/CharniakJ05} & C  & - & - & - &  92.1 & - & - & - & - \\
        \textbf{Ours (multi-branching)} & O$\uparrow$ (1) & \underline{1122.6} & 92.1 & 92.1 & 92.1 & O$\uparrow$ (1) & 86.0 & 84.7 & 85.3 \\
        \textbf{Ours (binary)} & O$\uparrow$ (1) & \textbf{1327.2} & \textbf{92.8} & \textbf{92.3} & \textbf{92.5} & O$\uparrow$ (1) & 85.8 & 86.2 & 86.0 \\
        \newcite{DBLP:conf/acl/NguyenNJL20} & O$\downarrow$ (1) & 130.2 & 92.8 & 92.8 & 92.8 & - & - & - & - \\
        \newcite{DBLP:conf/acl/KleinK18} & C & 212.5 & 93.9 & 93.2 & 93.6 & C & 91.9 & 91.5 & 91.7 \\
        \newcite{DBLP:conf/acl/WeiWL20} & O$\downarrow$ (1) & 155 & 94.1 & 93.3 & 93.7 & O$\downarrow$ (1) & 89.9 & 87.4 & 88.7 \\
        \newcite{DBLP:conf/acl/ZhouZ19} & C & 226.3 & 93.9 & 93.6 & 93.7 & C & 92.3 & 92.0 & 92.2 \\
        \newcite{DBLP:conf/ijcai/ZhangZL20} & C & \underline{1092} & 94.2 & 94.0 & 94.1 & C & 89.7 & 89.9 & 89.8 \\
        \hline
        \end{tabular}
        }
    \end{center}
    \caption{Single-model results on PTB and CTB test datasets sorted by the F1 scores on PTB. 
    Transition-based parsers, chart parser, and others are marked as T, C, and O, respectively;  
    $\uparrow$ and $\downarrow$ denote bottom-up and top-down. 
    The number in brackets indicates the beam size. 
    \newcite{DBLP:conf/acl/KleinK18} used Tesla K80, and the CTB scores are cited from \newcite{DBLP:conf/acl/KitaevCK19}.
    \newcite{DBLP:conf/acl/ZhouZ19} used GeForce GTX 1080 Ti (same condition).\label{tab:ovr}}
\end{table*}

Table \ref{tab:ovr} lists the parsing accuracies and speeds of the single models
in ascending order according to their F1 scores for the PTB corpus.
The transition-based parsers with $O(n)$ complexity appear at the top of the table,
followed by other types of models, and the chart parsers running in $O(n^3)$ time are at the bottom of the table.
The models exhibited similar trends for the CTB.
\newcite{DBLP:conf/acl/BengioSCJLS18} and our models belong to type O and have similar complexities.
Generally, the accuracy follows the complexity, whereas the speed roughly follows the year of publication rather than complexity or type.

\subsection{Comparison of Models} \label{sec:abl}
% \subsection{Use of Pre-Trained Language Model} \label{sec:pre}

\begin{table}[t]
    \begin{center}
        \scalebox{.9}{
        \begin{tabular}{c p{4.7cm} c}
        \hline
        \textbf{Var} & \centering \textbf{Specification} & \textbf{F1} \\
        \hline
        \textbf{B}/e & without fastText initialization. & 91.73 \\
        \textbf{B}/$\epsilon$ & with tuned official fastText. & 91.69 \\
        \textbf{B}/E & with frozen fastText from PTB. & 92.31 \\
        \hline
        % \textbf{B}/0 & without \texttt{BiLSTM}$_{cxt}$.      & 65.02 \\
        % \textbf{B}/1 & with 1-layer \texttt{BiLSTM}$_{cxt}$. & 89.27 \\
        % \textbf{B}/4 & with 4-layer \texttt{BiLSTM}$_{cxt}$. & 92.15 \\
        % \multicolumn{3}{c}{(See \textbf{frozen fastText} column in Table \ref{tab:dy_ctx}.)} \\
        % \hline
        \textbf{B}/F & \texttt{BiLSTM}$_{ori}$ into \texttt{FFNN}$^\star_{ori}$.   & 88.97 \\
        % \textbf{CV}/Q & \texttt{BiLSTM}$_{ori}$ into \texttt{BiQRNN}$_{ori}$. & \underline{92.42} \\
        \textbf{B}/L & \texttt{BiLSTM}$_{ori}$ with \texttt{BCE-LOSS}. & 92.32 \\
        \hline
        \end{tabular}
        }
    \end{center}
    \caption{Results of ablation studies on fastText (top) and \texttt{BiLSTM}$_{ori}$ (bottom)
    of the binary model.\label{tab:abl}}
\end{table}

\paragraph{Models with fastText.}
We investigated the binary model through ablation.
The impacts of fastText are presented in the upper part of Table \ref{tab:abl}.
\textbf{B}/E does not require any external data beyond PTB, which is comparable to models without a pre-trained GloVe \cite{DBLP:conf/emnlp/PenningtonSM14}.
% We trained fastText on PTB text and froze to train \textbf{B}/E.
% Models tuned with XLNet yield scores up to 95.7, whereas those tuned with fastText or trained without fastText yield scores that drop to 91.7.

Then, we replaced \texttt{BiLSTM}$_{ori}$ with an FFNN to examine its effect.
The results are in the bottom rows.
The comparison proves whether the embeddings are collaborative for the orientation signals
because FFNN regards each input independently.

\begin{figure}[t]
    \centering
    \includegraphics[width=\linewidth]{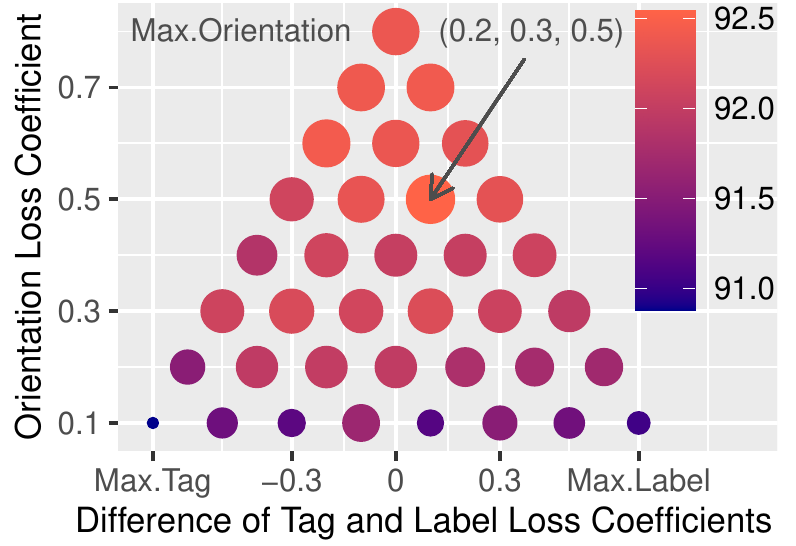}
    \caption{Grid search with an interval of 0.1 in a space of (tag, label, orientation) loss coefficients.
    The best was (0.2, 0.3, 0.5) indicated by an arrow. \label{fig:loss_weight}}
\end{figure}

Finally, we used a grid search to explore the hyperparameter space of our three-loss coefficients.
Figure \ref{fig:loss_weight} shows that the performance correlates to the orientation loss the most, 
but it is not overly sensitive to the hyperparameters. 

\begin{table}[t]
    \begin{center}
        \scalebox{.9}{
        \begin{tabular}{c | c c | c c}
        \hline
        & \multicolumn{2}{c |}{\textbf{Frozen fastText}} & \multicolumn{2}{c}{\textbf{Frozen XLNet}} \\
        \textbf{Var} & \textbf{F1} & \textbf{sents/sec} & \textbf{F1} & \textbf{sents/sec}\\
        \hline
        \textbf{B}/0 & 65.02 & \underline{1386.6} & 89.24 & \underline{411.2} \\
        \textbf{B}/2 & 91.34 & 1350.0 & 93.74 & 398.4 \\
        \textbf{B}/6 & \underline{92.54} & 1327.2 & \underline{93.89} & 382.7 \\
        \hline
        \end{tabular}
        }
    \end{center}
    \caption{Effectiveness of using frozen static word embeddings or dynamic sub-word language model and corresponding peak speed. \label{tab:dy_ctx}}
\end{table}

\paragraph{Pre-trained language model.}
We compared the results using frozen fastText with those using frozen XLNet\footnote{
    XLNet tokenizes words into sub-word fractions.
    For the frozen XLNet, using leftmost, rightmost, or averaged sub-word embeddings as the word input yielded similar results.
    
} in Table~\ref{tab:dy_ctx}.
The accuracy of the model increased along with the depth of \texttt{BiLSTM}$_{cxt}$, and
it exhibited the most significant increase across all variants.
Owing to XLNet, our complexities grew to $O(n^2)$.

\begin{table}[t]
    \begin{center}
        \scalebox{.9}{
        \begin{tabular}{l | c c c}
        \hline
        \textbf{Fine-Tuned Model} & \textbf{F1} & \textbf{sents/sec} & \textbf{Type} \\
        \hline
        \newcite{DBLP:conf/acl/KleinK18} & 95.13 & 70.8  & C \\
        \newcite{DBLP:conf/acl/KitaevK20} & 95.44 & \underline{1200}  & T \\
        \newcite{DBLP:conf/acl/NguyenNJL20} & 95.48 & - & O$\downarrow$ \\
        \newcite{DBLP:conf/ijcai/ZhangZL20} & 95.69 & - & C \\
        \newcite{DBLP:conf/acl/WeiWL20} & 95.8 & - & O$\downarrow$ \\
        \newcite{DBLP:conf/acl/ZhouZ19} & 96.33 & 64.8 & C \\
        \newcite{DBLP:conf/nips/YangD20} & 96.34 & 71.3 & T \\
        \newcite{DBLP:conf/emnlp/MriniDTBCN20} & \underline{96.38} & 59.2 & C \\
        \hline
        \textbf{B}/0 (XLNet+FFNN) & \underline{95.72} & \underline{411.2}  & O$\uparrow$ \\
        \textbf{B}/2 (XLNet+BiLSTM) & 94.67 & 398.4  & O$\uparrow$ \\
        \textbf{M}/0 (XLNet+FFNN) & 95.44 & 369.4 & O$\uparrow$ \\
        \hline
        \end{tabular}
        }
    \end{center}
    \caption{Improvements with pre-trained language models.
    We used a greedy search algorithm on single GeForce GTX 1080 Ti.
    Rows 6--8 are reported by \newcite{DBLP:conf/nips/YangD20} using GeForce GTX 2080 Ti.
    \newcite{DBLP:conf/acl/KitaevK20} used a cloud TPU with a beam search algorithm and a larger batch size.
    \label{tab:tune}}
\end{table}

We fine-tuned our models\footnote{
    For the fine-tuned XLNet, 
    using either the leftmost or rightmost sub-word yielded similar results earlier.
    However, averaging sub-words produced F1 scores under 94.
} and compared them with other parsers using fine-tuned language models. These are listed in Table~\ref{tab:tune}.

\subsection{Tree-Binarization Strategy} \label{sec:lin}

\begin{figure}[t]
    \centering
    \includegraphics[width=\linewidth]{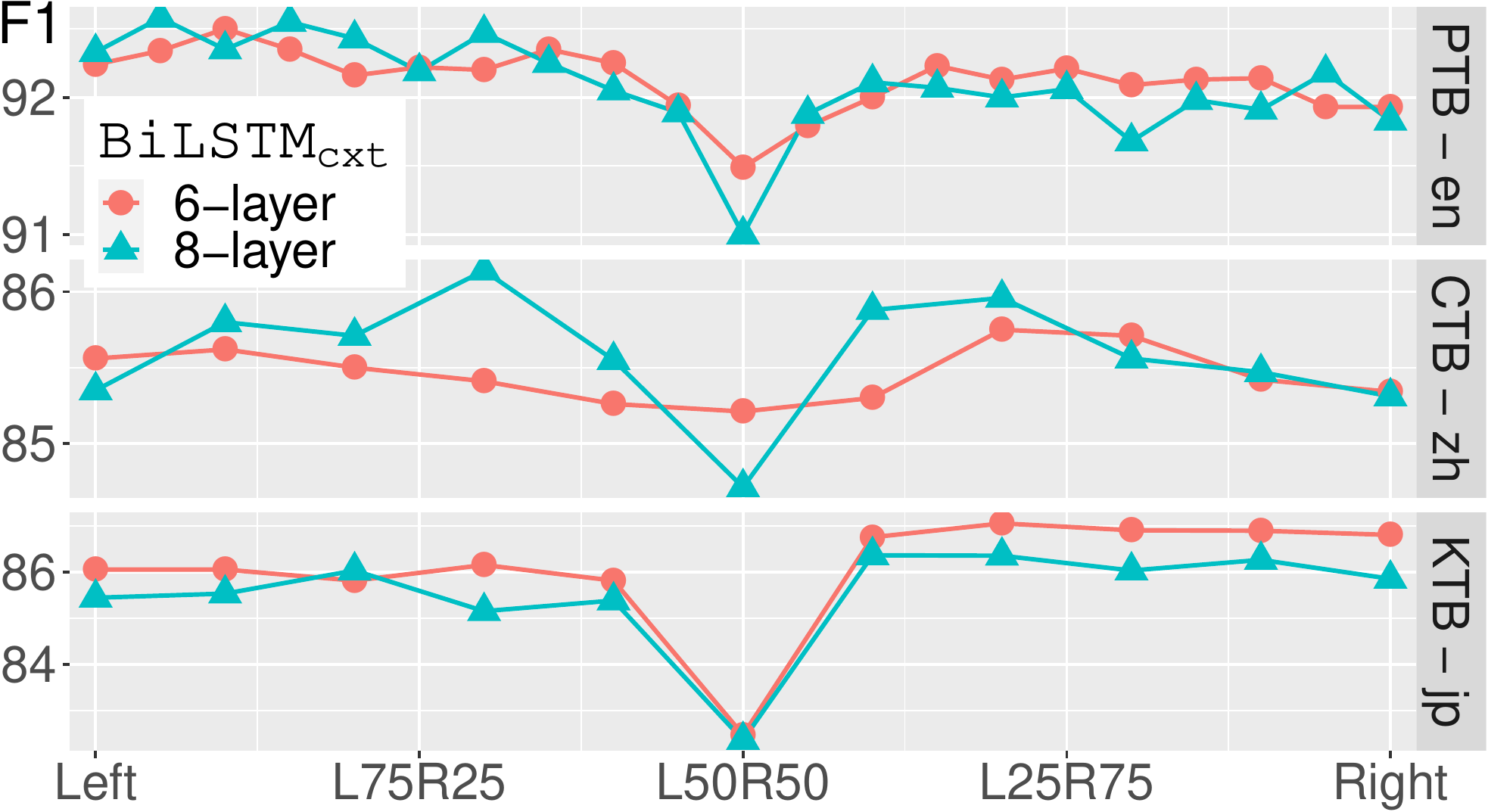}
    \caption{Probabilistic interpolations of two CNF factors to F1 scores.
    The capacity of \texttt{BiLSTM}$_{cxt}$ is almost saturated with 6 or 8 layers.
    \label{fig:cnf}}
\end{figure}

To reflect the branching tendency, our best single model for PTB was obtained on the dynamic L95R05 dataset. This dataset is 
a probabilistic interpolation between the left-factored dataset (for 95\% chances) 
and a right-factored dataset (for 5\% chances) in Figure \ref{fig:cnf}.
The best model for CTB appeared on the left side at L70R30, scoring $86.14$, whereas
the best for KTB was on the L30R70 dataset, scoring $87.05$ with a 6-layer \texttt{BiLSTM}$_{cxt}$.
Typically, the results for all the corpora had a minimum at L50R50.
For English, the left ``wing'' was higher than the right; the opposite trend was observed for Japanese.
For Chinese, no clear trend was obtained.

All studies described in the previous sections were conducted on the PTB L85R15 dataset.
% because we frequently obtained the best models from L95R05 to L75R25.

\subsection{Complexity and Speed} \label{sec:cpl_spd}
\begin{figure}[t]
    \centering
    \includegraphics[width=.7\linewidth]{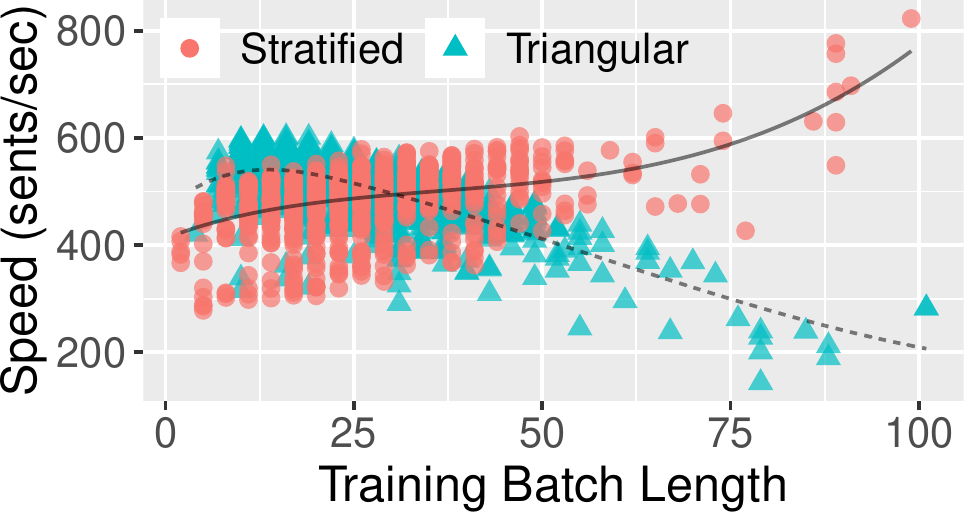}
    \includegraphics[width=.288\linewidth]{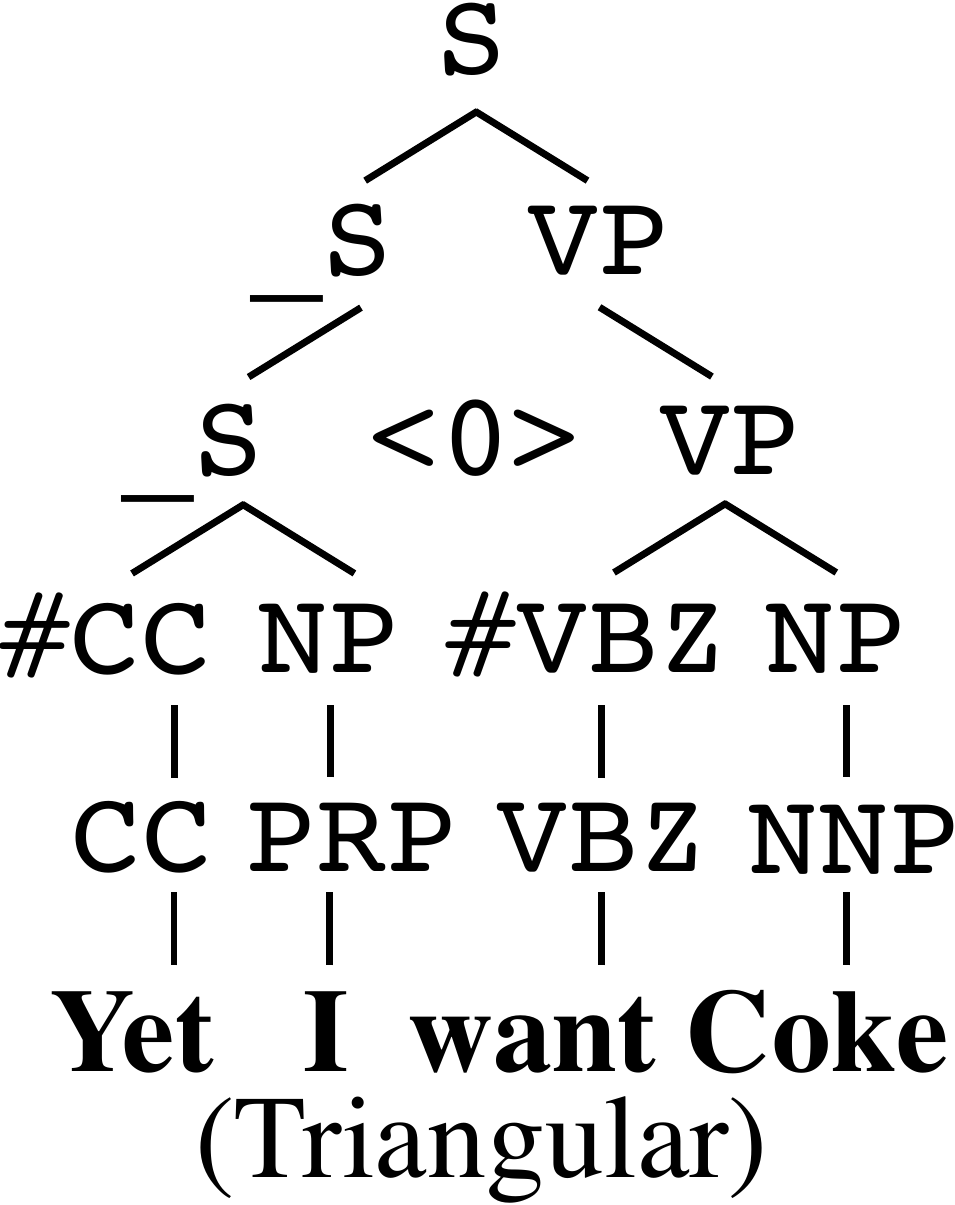}
    \caption{Linear complexity vs. squared complexity. 
    Redundancy with placeholder ``\texttt{<0>}'' helps maintain the triangular shape. \label{fig:speed}}
\end{figure}

\begin{table}[t]
    \begin{center}
        \scalebox{.9}{
        \begin{tabular}{c | c c c}
        \hline
        \textbf{Format} & \textbf{Time/150} & \textbf{Memory} & \textbf{OOM} \\
        \hline
        \textbf{Stratified} & 7.5 hours & 3.3 GB & - \\
        \textbf{Triangular} & 15.9 hours & 8.2 GB & 100 \\
        \hline
        \end{tabular}
        }
    \end{center}
    \caption{Training time and memory consumed by our two data formats.
    The time column indicates the time used for 150 training epochs with validations.
    Development F1 scores are approximately 92.4.
    The OOM column lists the length limit for preventing an out-of-memory error.
    \newcite{DBLP:conf/acl/KleinK18} took 10 hours for 93 training epochs on our GeForce GTX 1080 Ti to yield their results. \label{tab:fmt}}
\end{table}

To test our linear speed advantage,
we inflated our training data with redundant nodes to resemble the triangular chart of CYK algorithm,
as depicted in Figure~\ref{fig:speed} and Table \ref{tab:fmt}.
The parse in the triangular treebank has the worst-case complexity of $O(n^2)$.
Meanwhile, training with linearity halved the training time,
reduced memory usage, and canceled the length limit for our three corpora.
There is a sheer difference between linearity and squared complexity.

\section{Discussion}

\subsection{Model Structure}

Our parsers comprise a neural encoder for scoring (i.e., Algorithm \ref{alg:enc}) and a non-neural decoder for searching. 
% and assembling the symbolic tree (See Appendix \ref{sec:symbolic}).
The decoder is a symbolic extension of the encoder in that both run in bottom-up manner, 
and the decoder interprets the scores as local-and-greedy decisions.
Other neural parsers also fit a similar encoder--decoder framework.
However, decoders with dynamic programming often include forward and backward processes
heterogeneous to their forward encoders~\cite{DBLP:conf/acl/KleinK18,DBLP:conf/acl/KitaevK20}.
The encoder and decoder in our model and \newcite{DBLP:conf/acl/BengioSCJLS18} are more homogeneous and can be easily merged.
Our parsers are bottom-up combinatory, while theirs was top-down splitting.
Similar homogeneity can be found in an easy-first dependency parser \cite{DBLP:conf/naacl/GoldbergE10}.

\begin{figure}[t]
    \centering
    \includegraphics[width=\linewidth]{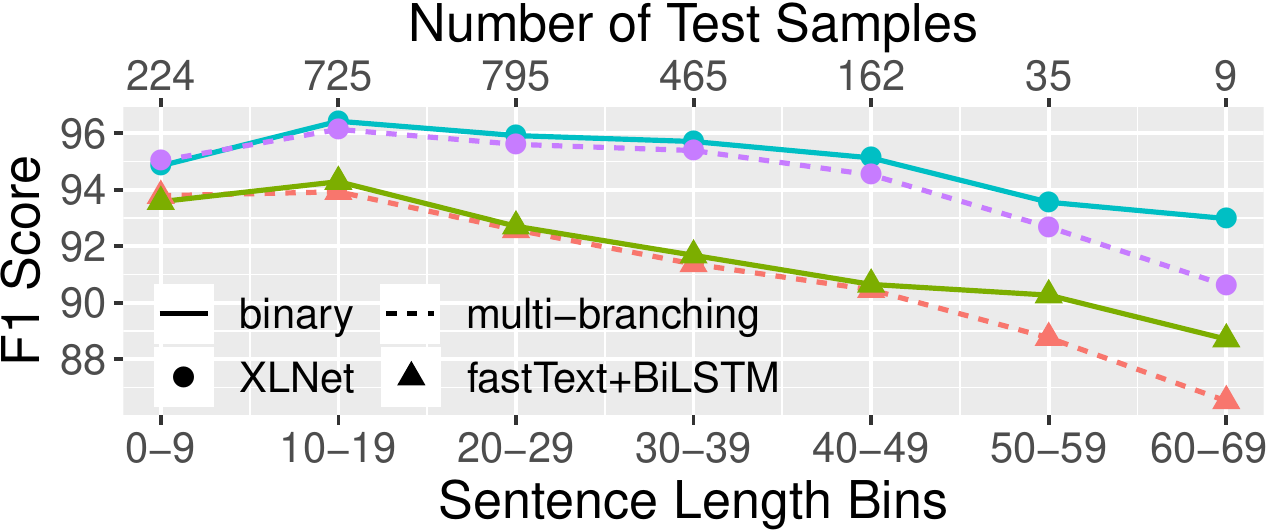}
    \caption{XLNet provides an overall improvement for all models and length bins.
    % ``\textbackslash f'' and ``\textbackslash t'' stand for \underline{f}rozen and \underline{t}uned.
    All models find it challenging to handle long sentences.
    \label{fig:score}}
\end{figure}

\paragraph{Input component.}
In terms of encoder, Tables \ref{tab:abl}--\ref{tab:tune} 
examine the impact of \texttt{BiLSTM}$_{cxt}$ with fastText or XLNet,
and the following conclusions can be drawn.
% 1) If the pre-trained model is frozen, input contextualization or transformation with \texttt{BiLSTM}$_{cxt}$ is necessary.
% The performances significantly degrade when \texttt{BiLSTM}$_{cxt}$ is absent,
% which contains up to 4.3M out of the 4.6M model parameters. Further, its recurrent dropout of 0.2
% accounts for approximately 0.4 points of POS tagging accuracy and 0.5 points of the F1 score.
% If the pre-trained model is tuned, opposite trends are obtained for fastText and XLNet.
1) The top rows of Table \ref{tab:abl} suggest that frozen fastText embeddings contain sub-word information,
whereas tuning them disturbs the frozen information because the n-gram model is not part of our model.
2) Table \ref{tab:dy_ctx} shows that the deeper the contextualization \texttt{BiLSTM}$_{cxt}$ (or XLNet), the better the results.
3) Tables \ref{tab:dy_ctx} and \ref{tab:tune} indicate that the tuning process
for the pre-trained language models~\cite{DBLP:conf/naacl/PetersNIGCLZ18,DBLP:conf/naacl/DevlinCLT19,DBLP:journals/corr/abs-1906-08237} achieves a significant improvement.

\paragraph{Speed and size.}
One of our research goals was to achieve simplicity and efficiency.
In terms of speed, our models parallelize more actions than transition-based parsers and have fewer computing nodes than chart parsers.
In terms of size, our models contain approximately 4M parameters in addition to the 13M fastText (or 114M XLNet) pre-trained embeddings,
which is fewer than those of \newcite[22M+]{DBLP:conf/acl/BengioSCJLS18} and \newcite[26M]{DBLP:conf/acl/KleinK18}.
% Meanwhile, our models are comparable to the aforementioned models in terms of accuracy while maintaining considerably high speeds.
The recursiveness and productiveness of vector compositionality should account for the compact size.

\paragraph{Vector compositionality.}
The performance of \texttt{FFNN}$_{ori}^\star$ is inferior to that of its RNN counterparts, suggesting that
some information might not be encoded locally. Thus, 
the \texttt{COMPOSE} function should remain in a contextual form to collaboratively leverage the whole layer.
However, BiRNN might still be a bottleneck for long-range orientation, as suggested in Figure \ref{fig:score}.
\texttt{BiLSTM}$_{chk}$ is a major weakness of the multi-branching model, especially for longer sentences.
% We have not yet found a better solution for it.
% The recurrent signal seems to vanish regardless of the model type, whereas short-range inference advantages are evident.

\subsection{Tree Binarization and Headedness}

\paragraph{Tree binarization.} 
Probabilistic interpolation with two CNF-factored datasets is effective for the three languages studied, as shown in Figure \ref{fig:cnf}.
Dynamic sampling allows the model to cover a wider range of composed vectors to improve its robustness to ambiguous orientations.
Furthermore, it seems counterintuitive for human learners to obtain the best model
using left-biased interpolation for a right-branching language 
or vice versa.
% or using right-biased interpolation for a left-branching language. 
However, for a neural model, balancing the frequency seems to be the key factor for improving performance 
\cite{DBLP:conf/acl/SennrichHB16a,DBLP:conf/emnlp/ZhaoZHZ018}.
The fact that the L50R50 dataset yielded the worst models also suggests that
the balance should be based on a default orientation tendency.
This could also be the reason why \textbf{mid-in} or \textbf{mid-out} did not improve the model.

\begin{figure}[t]
    \centering
    \includegraphics[width=\linewidth]{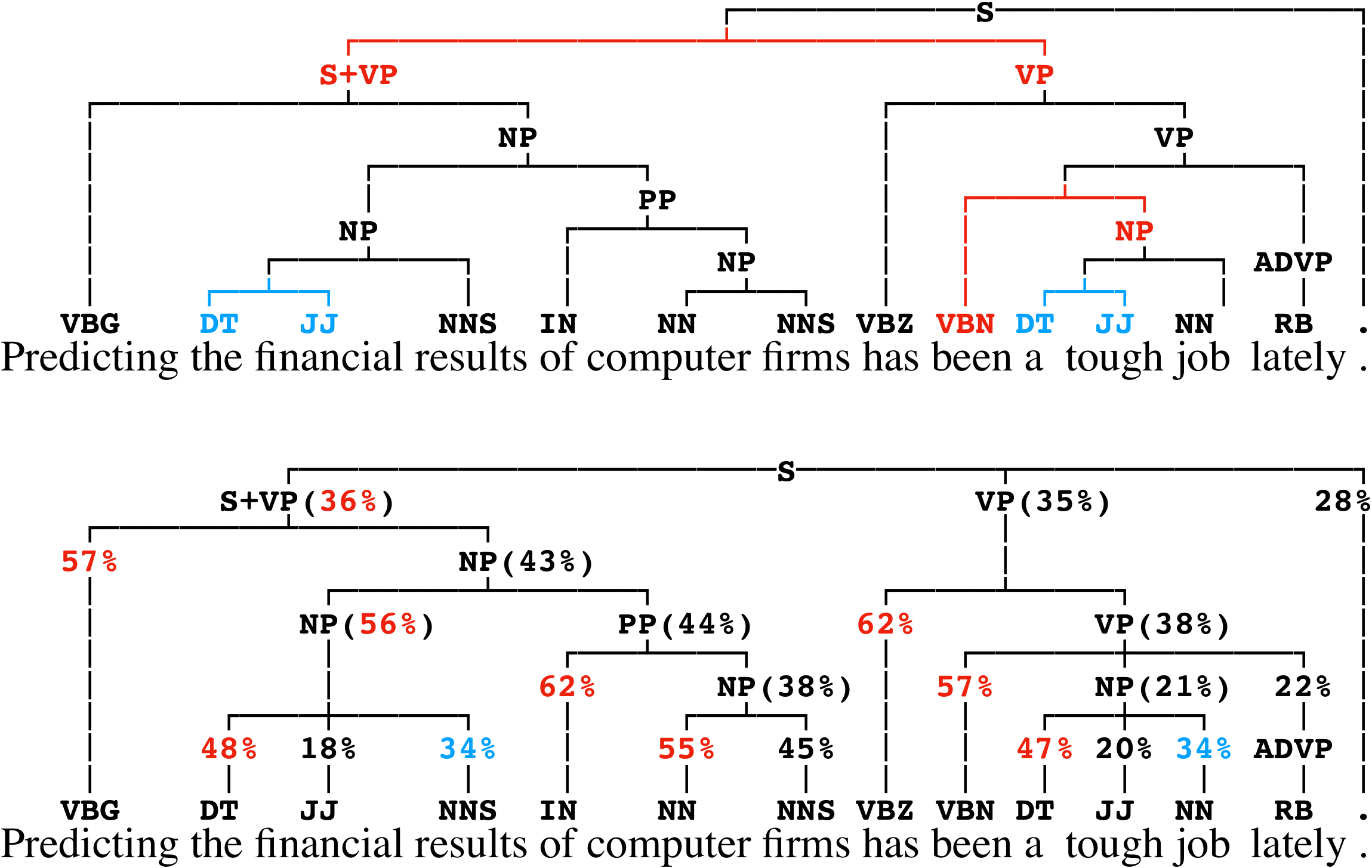}
    \caption{English internal constituents (top) and headedness (bottom) from our two models.
    \label{fig:m_ary}}
\end{figure}

\paragraph{Headedness.}
Figure \ref{fig:m_ary} show the intermediate parses on the same sentence from our two models.
They are typical examples in the output.

The binary model first combines determiners and their right neighbors rather than adjectives and nouns in noun phrases (blue spans).
It also postpones the combination with adjuncts such as punctuation and adverb (red spans).
The high frequencies of determiners in noun phrases make them great attractors.
% Punctuation and adverb are the opposite; greater attractors combine first. 

\begin{table}[t]
    \setlength\tabcolsep{5pt}
    \begin{center}
        \scalebox{.9}{
        \begin{tabular}{l p{5.72cm}}
        \hline
        \hline
        \textbf{Parent} (\#) & \textbf{Head child by maximum weight} \\
        \hline
        \hline
        \texttt{NP}\hspace*{\fill}(14.4K) & \texttt{DT} (4.5K); \texttt{*NP} (4.3K); \texttt{*NNP} (1.6K); \texttt{*JJ} (922); \texttt{*NN} (751); \texttt{*NNS} (616); etc. (1.6K; 38 of 50 types with ``\texttt{*}'') \\
        \hline
        \texttt{VP}\hspace*{\fill}(6.8K) & \texttt{VBD} (1.5K); \texttt{VB} (1.4K); \texttt{VBZ} (1.0K); \texttt{VBN} (954); \texttt{VBP} (705); \texttt{MD} (523); \texttt{VBG} (387); \texttt{VP} (169); \texttt{TO} (81); etc.\\ %  (24; 10 types) 
        \hline
        \texttt{PP}\hspace*{\fill}(5.5K) & \texttt{IN} (5.0K); \texttt{TO} (397); etc. \\ % (148; 14) 
        \hline
        \texttt{S}\hspace*{\fill}(3.8K) & \texttt{VP} (3.4K); \texttt{S} (194); \texttt{NP} (90); etc. \\
        \hline
        \texttt{SBAR} \hspace*{\fill}(1.2K) & \texttt{IN} (649); \texttt{WHNP} (395); \texttt{WHPP} (19); \texttt{WHADVP} (121); \texttt{SBAR} (15); etc. \\
        \hline
        \texttt{ADVP}\hspace*{\fill}(278) & \texttt{RB} (181); \texttt{IN} (30); \texttt{RBR} (25); etc. \\
        \hline
        \texttt{QP}\hspace*{\fill}(198) & \texttt{CD} (67); \texttt{IN} (65); \texttt{RB} (29); \texttt{JJR} (16); \\
        \hline
        \end{tabular}
        }
    \end{center}
    \caption{English headedness selection with our multi-branching model on PTB test set.  
    ``\texttt{*}'' marks the absence of a \texttt{DT} child for its \texttt{NP} sisters. 
    For quantifier phrases (\texttt{QP}), some non-quantifiers are more likely to be heads if they appear;
    e.g., adverbs (\texttt{RB}; e.g., ``approximately''),
    prepositions (\texttt{IN}; e.g., ``about''),
    and relative adjectives (\texttt{JJR}; e.g., ``more than''). \label{tab:head}}
\end{table}

On the other hand, the multi-branching model places close attention on what the syntactic head is supposed to be.
In the noun phrases, determiners receive the highest weight averages (red), and the nouns obtain the second (blue).
This phenomenon suggests that an English noun phrase's syntactic role is mainly projected from the determiners, as discussed by \newcite{zwicky_1985}.
Table \ref{tab:head} provides more statistical support.
For example, the model selects \texttt{DT} as an \texttt{NP} head if it is available; otherwise, nouns and adjectives are prominent heads.
Chinese and Japanese parsers work similarly for their headedness.
(See Appendix \ref{sec:zhjp}.)
% For example, English determiners may better cast their syntactic roles to noun phrases than nouns.

\subsection{Error Analysis}
\begin{figure}[t]
    \centering
    \includegraphics[width=.65\linewidth]{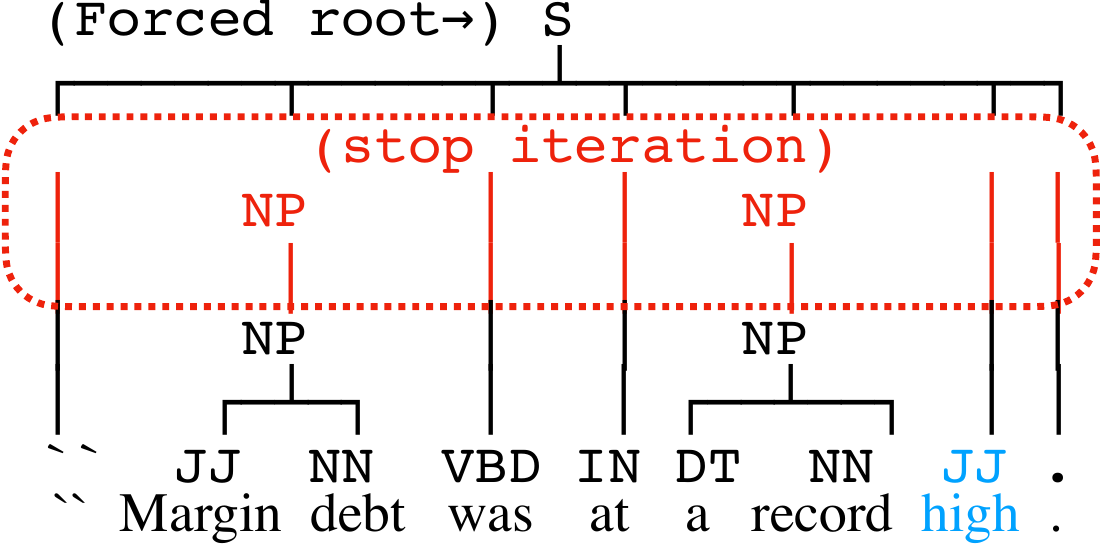}
    \caption{Failed parse from the multi-branching model.
    The model stops parsing and saves computations when it repeats the same chunking positions.
    \label{fig:m_ary_err}}
\end{figure}

The rate of an invalid parse is the last topic that we consider for our parsers.
For the binary parser, fatal errors, such as frame-breaking orientations, appear at an early stage of training.
However, the late $90\%$ of training time contains very few errors, and our binary model is free from invalid parsing on the test set.
For the multi-branching parser, it is observed that 11 out of 2,416 test parses are forests rather than parse trees when they are trained with fastText.
However, the multi-branching parser with fine-tuned XLNet reduces the error count on the test set to 1.
% (See Appendix \ref{sec:behavior}.)

We present a failed multi-branching parse with fastText, as shown in Figure \ref{fig:m_ary_err}.
The postnominal adjective ``high'' is uncommon for English.
Because the model did not group it with the adjacent ``a record'' to form an \texttt{NP},
the error propagated to higher layers (e.g., no \texttt{PP} as an adjunct to form a \texttt{VP}), causing the bad parse.
It implies that the multi-branching model requires an appropriate predict-argument configuration to chunk.

\section{Conclusion}
We proposed a pair of neural combinatory constituency parsers.
The binary one yields F1 scores comparable to those of recent neural parsers.
The multi-branching one reveals constituency headedness.
Both are simple and efficient with relatively high speeds.
We also leveraged a pre-trained language model and CNF factors
to increase the accuracy.
We reflected the branching tendencies of three languages.

\section*{Acknowledgments}
We extend special thanks to our reviewers for invaluable comments
and Hayahide Yamagishi for initial discussions. % the 
This work has been partly supported by the Grant-in-Aid for Scientific Research 
from the Japan Society for the Promotion of Science (JSPS KAKENHI); Grant Number 19K12099.

\bibliography{acl2021}
\bibliographystyle{acl_natbib}

\newpage
\appendix
\section{Appendices}

\subsection{Compression Ratios and Linearity} \label{sec:lang_ratio}

\begin{figure}[t]
    \centering
    \includegraphics[width=\linewidth]{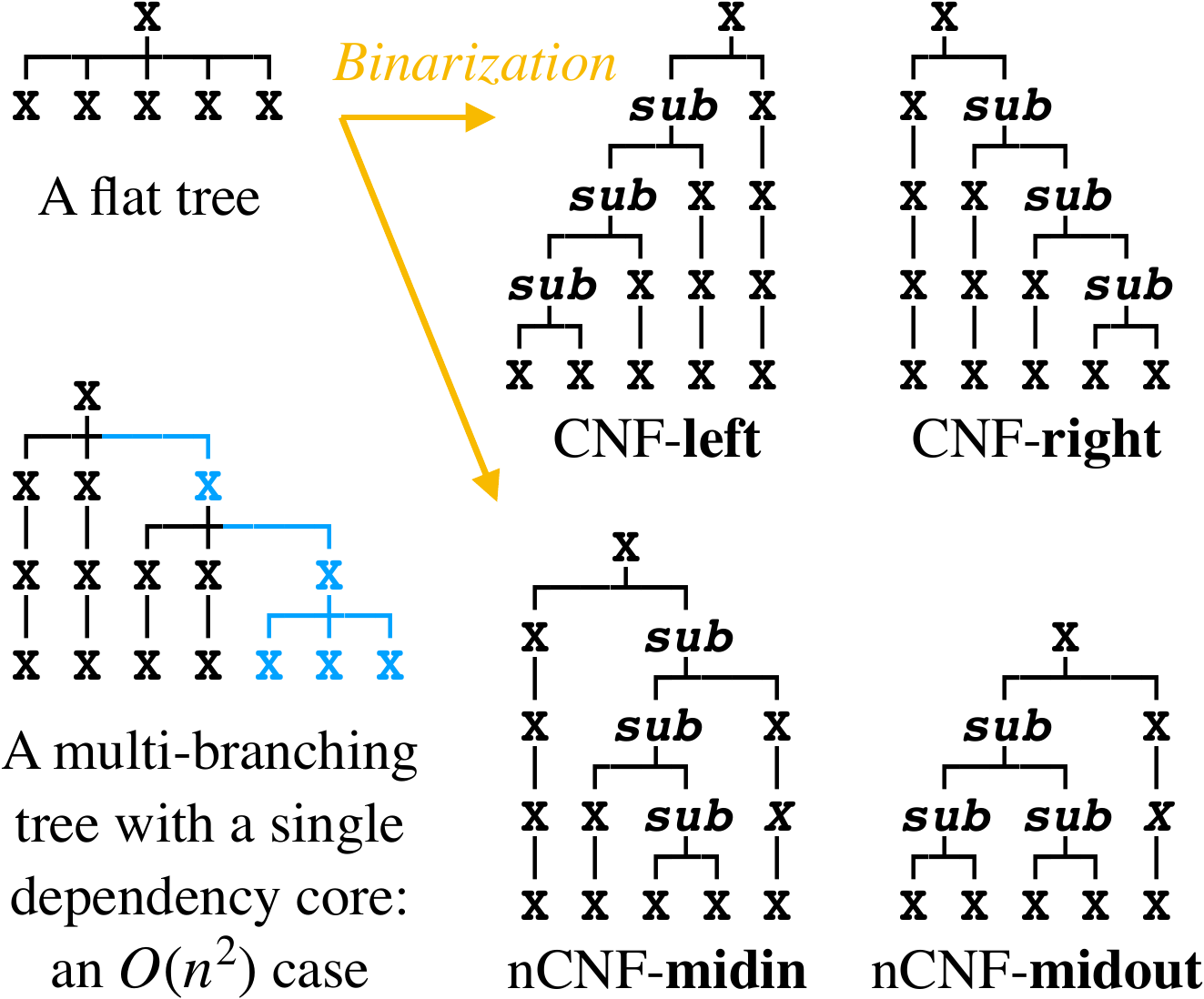}
    \caption{
        Examples used in the content. A flat tree is binarized with four factors.
        The binarization of CNF-\textbf{left}, CNF-\textbf{right}, or nCNF-\textbf{midin}
        creates binary trees with a single dependency core (i.e., a single \textit{sub} thread), whose $O(n^2)$ complexity is the same with the tree without binarization in the lower left.
        Most nodes are relaying nodes.
        Meanwhile, nCNF-\textbf{midout} enables concurrent phrases with $O(n)$ complexity (i.e., multiple \textit{sub} threads).
        However, the division tends to break a constituent into ungrammatical pieces, which confuses the model and does not lead to improvement.
    \label{fig:trees}}
\end{figure}

\begin{figure}[t]
    \centering
    \includegraphics[width=.78\linewidth]{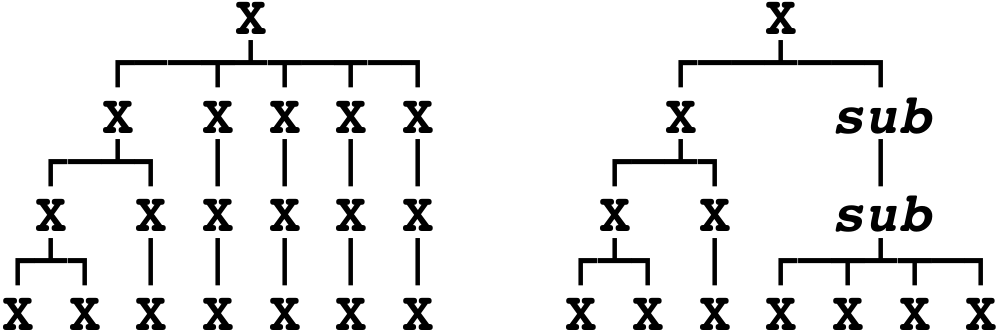}
    \caption{
        Adding \textit{sub} nodes to make flat structure more efficient.
        Using the strategy as a new dynamic dataset also brings multi-branching model \textbf{M} a stable accuracy improvement with an F1 score of 92.36 on PTB.
        However, it has nothing to do with linguistic properties. We save it for a future study.
    \label{fig:m_ary_sub}}
\end{figure}

\begin{figure*}[t]
  \centering
  \includegraphics[width=.5223\linewidth]{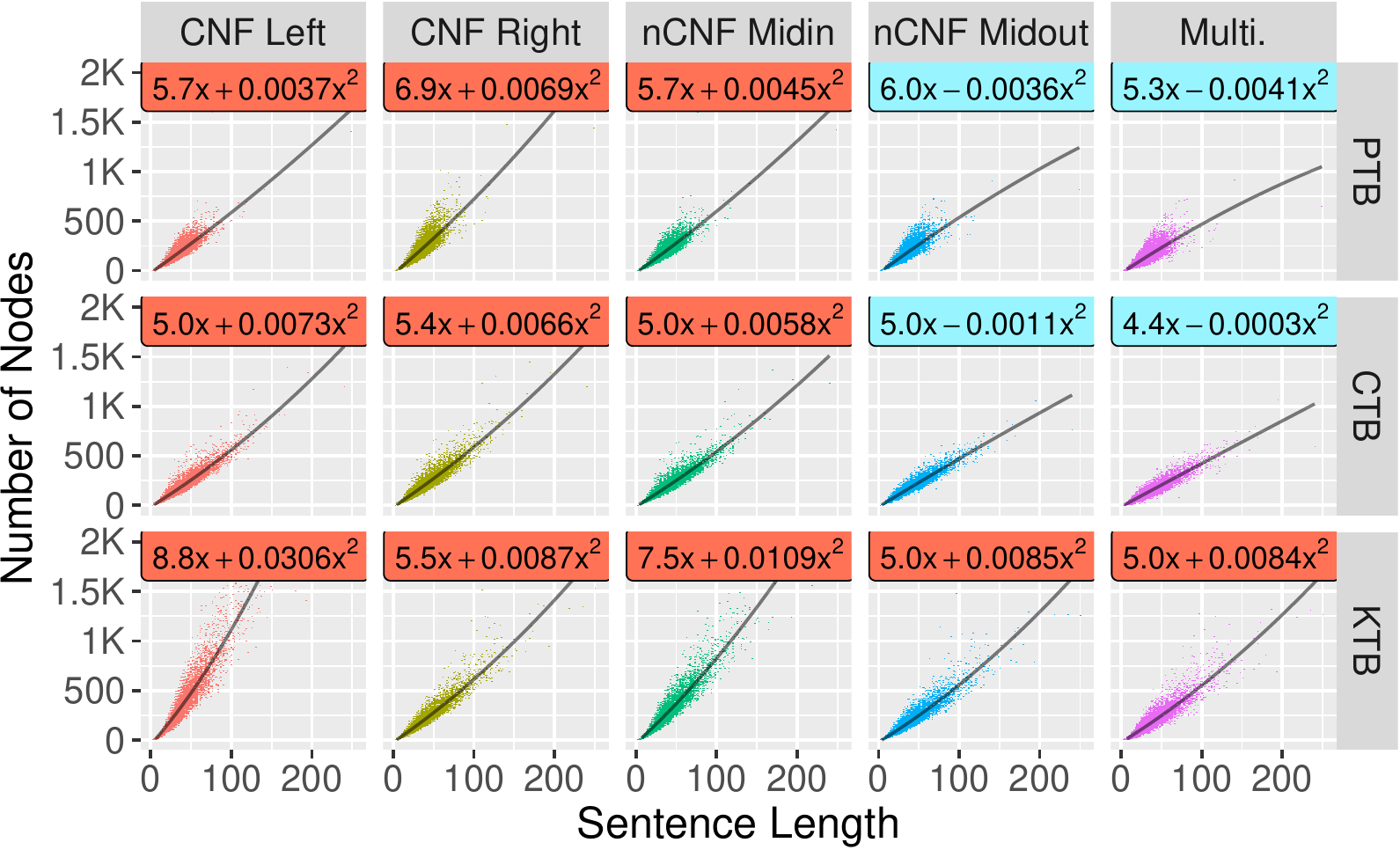}
  \includegraphics[clip, trim=1.62cm 0 0 0, width=.47\linewidth]{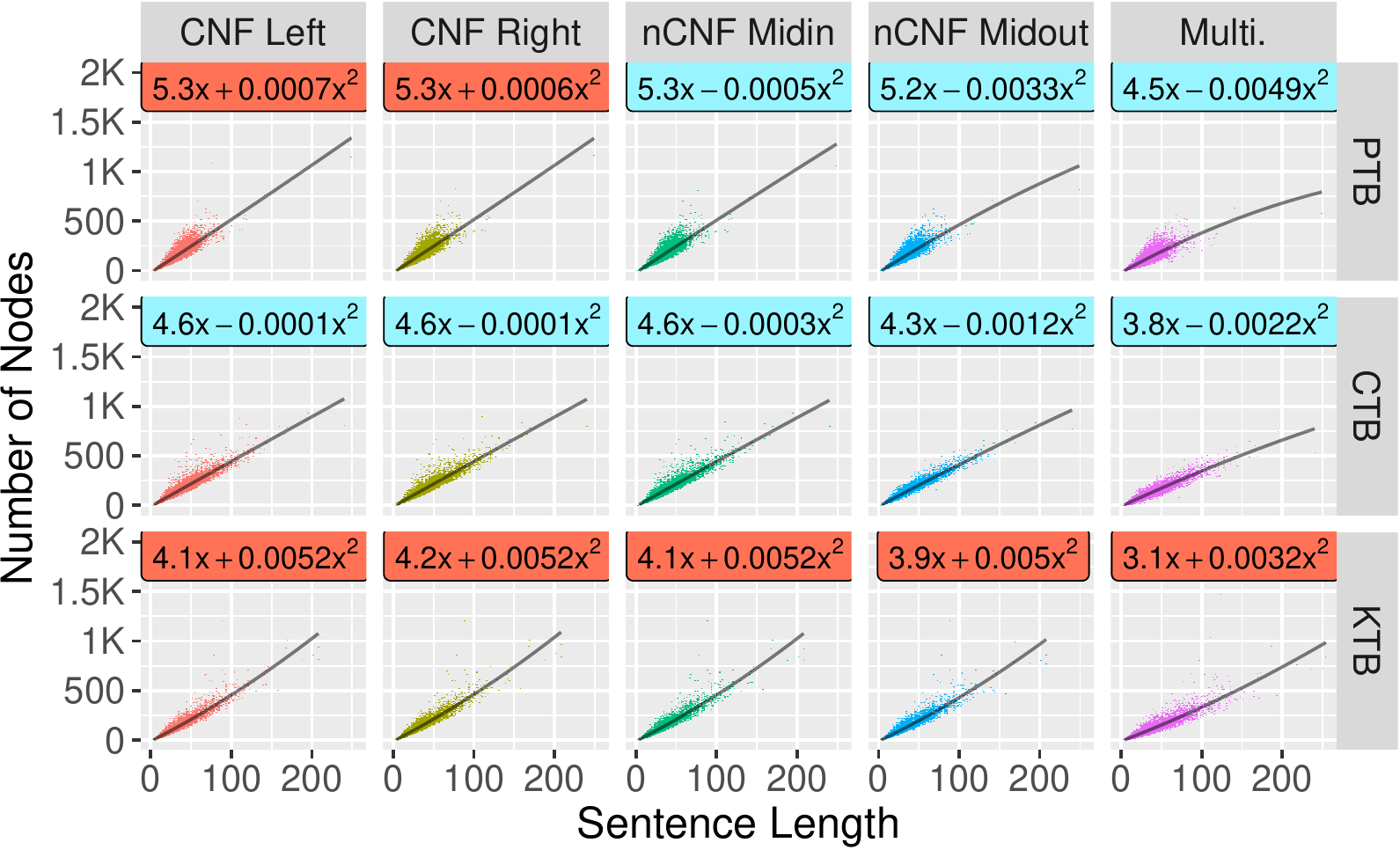}
  \caption{
      Left: the empirical complexities related to Figures \ref{fig:level_ratio_full} \& \ref{fig:level_ratio_multi_full}.
      Linear regressions are shown on a light blue background when the quadratic terms are negative.
      Right: resultant complexities after a preprocessing that groups the flat structure into sub-constituent before stratification.
      (See Figure \ref{fig:m_ary_sub}.)
  \label{fig:all_complexities}}
\end{figure*}

\begin{figure*}[t]
    \centering
    \includegraphics[width=.3505\linewidth]{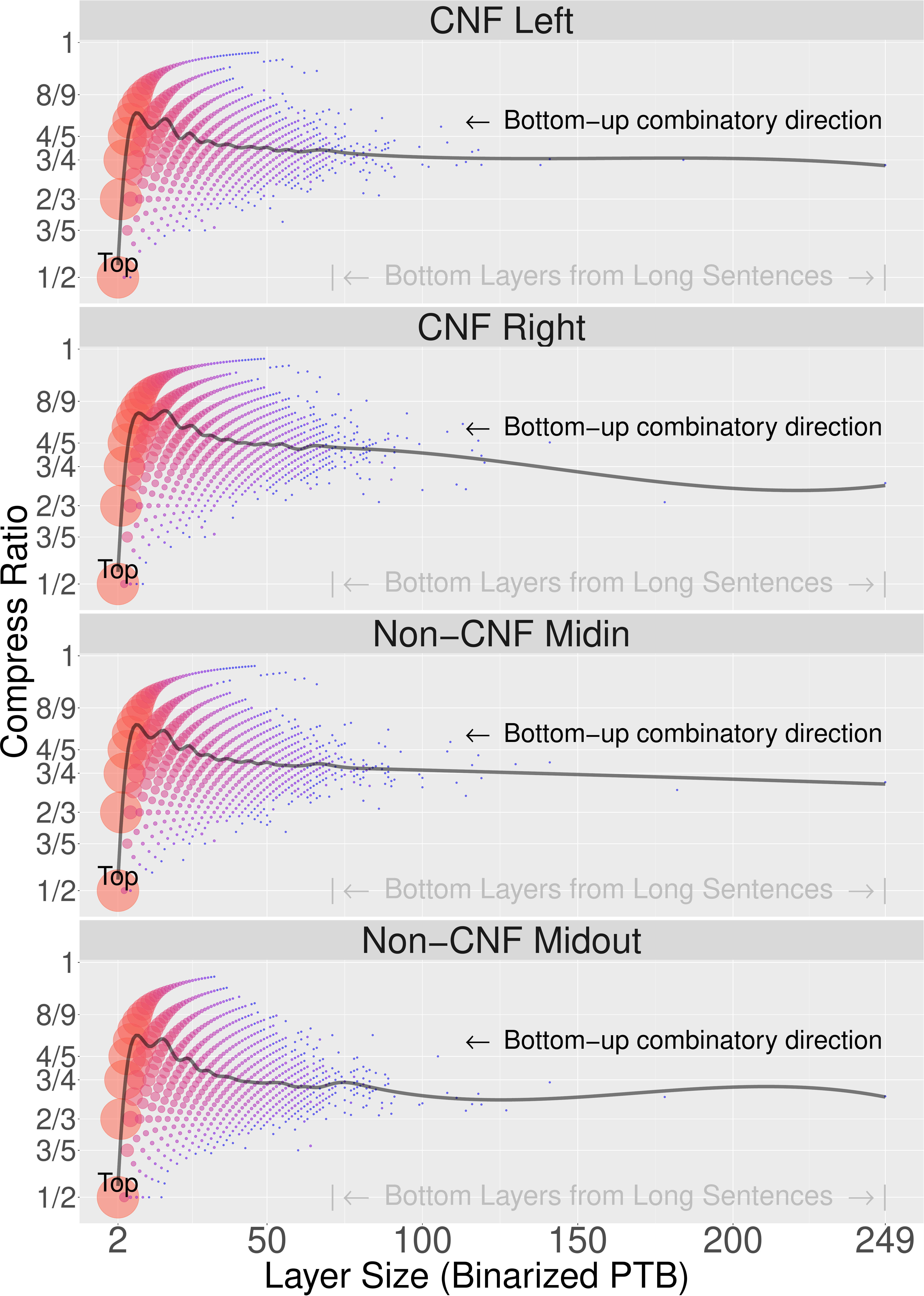}
    \includegraphics[clip, trim=4.5cm 0 0 0, width=.31925\linewidth]{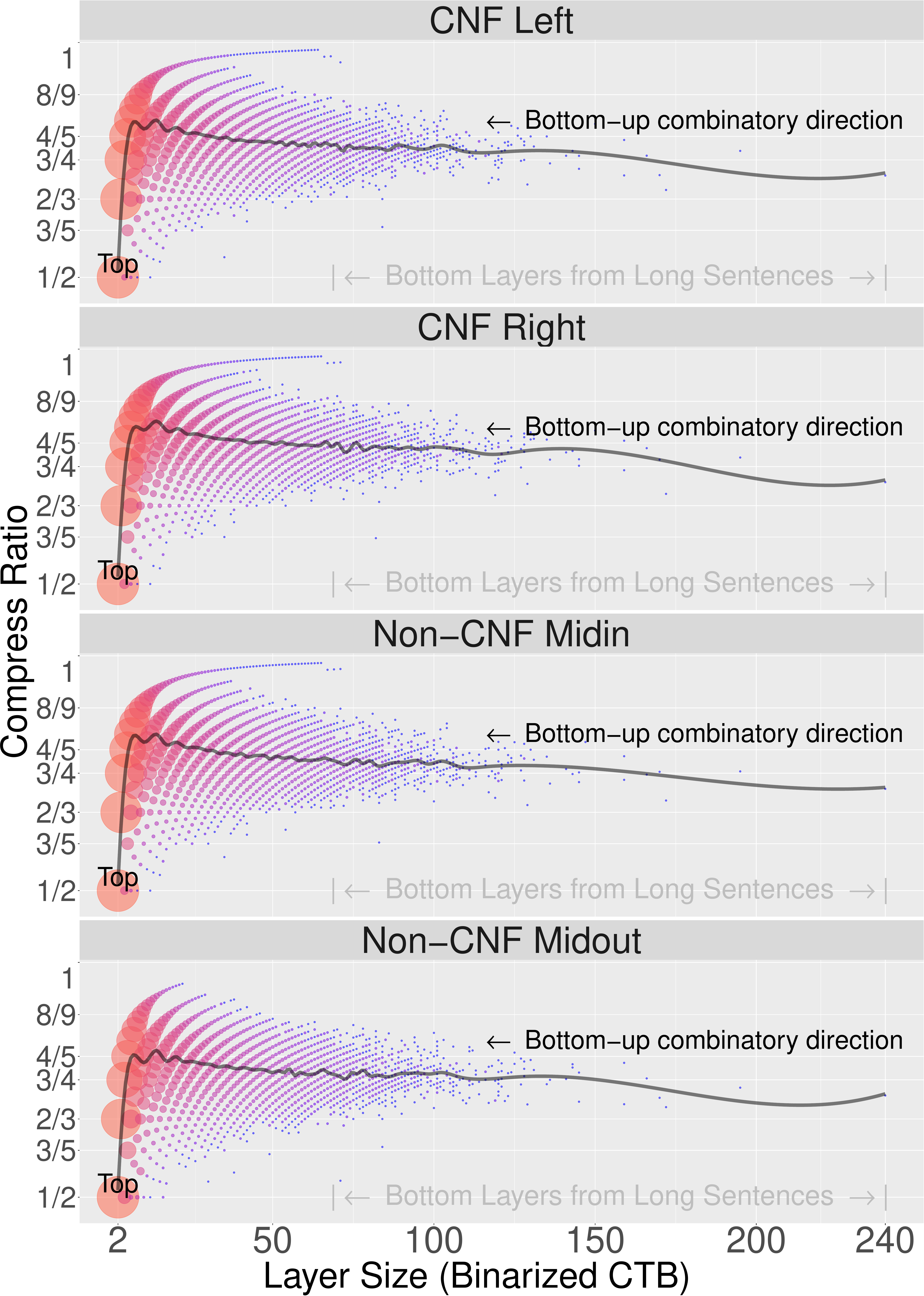}
    \includegraphics[clip, trim=4.5cm 0 0 0, width=.31925\linewidth]{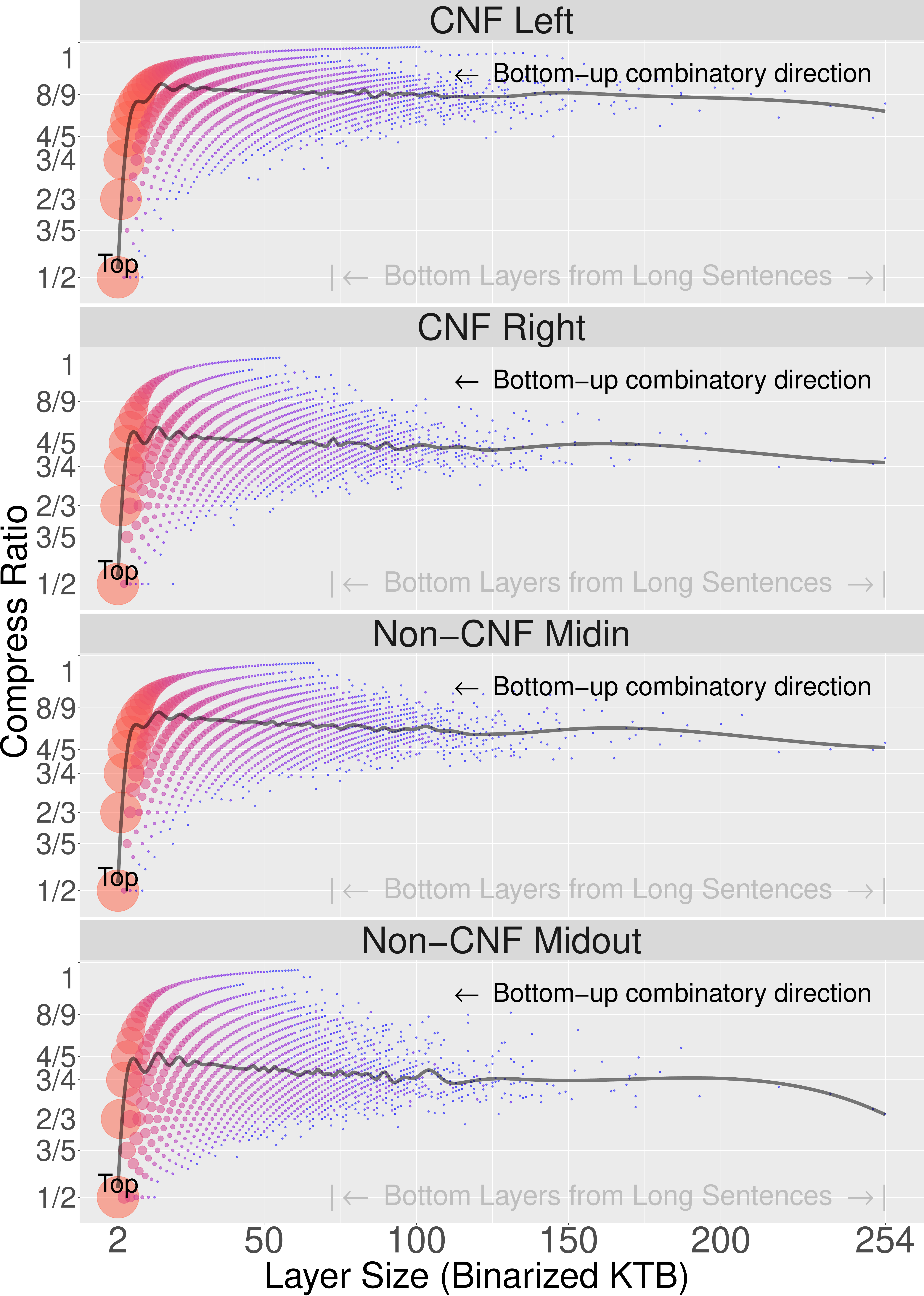}
    \caption{Binarized corpora with four factors. Curved tiers can be observed in each plot.
    For example, the leftmost tier is composed of $\frac{n-1}{n}$ (followed by $\frac{n-2}{n}$, $\frac{n-3}{n}$, and so on). 
    The dots in this tier range from a high compression ratio of 0.5 to the least efficient ones in their corpus.
    Efficient dots are more populated, judging by their sizes and colors.
    % Because the left side is closed and limited by their lengths, the most influential part is the right open side,
    All statistics yield stable means, which are also presented in Table \ref{tab:comp_ratios}.\label{fig:level_ratio_full}}
\end{figure*}

\begin{figure*}[t]
    \centering
    \includegraphics[width=\linewidth]{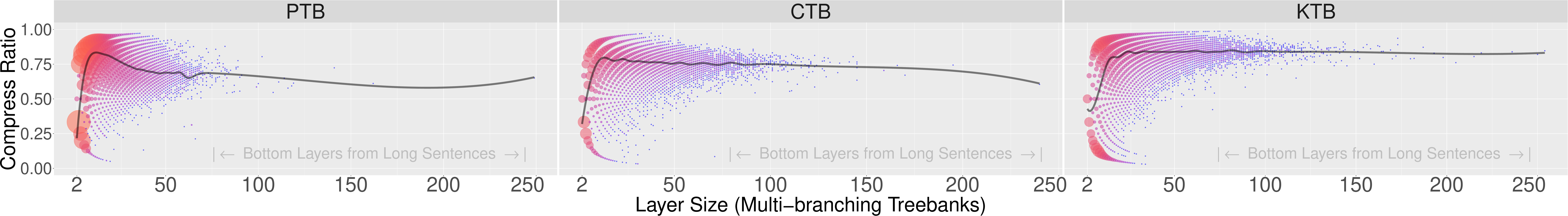}
    \caption{Multi-branching corpora. Curved tiers appear more symmetric and all statistics still yield stable means.\label{fig:level_ratio_multi_full}}
\end{figure*}

\begin{table*}[t]
    \begin{center}
    \scalebox{.9}{
    \begin{tabular}{c c c c c c}
    \hline
    \textbf{Factor} & \textbf{Left} & \textbf{Right} & \textbf{Midin} & \textbf{Midout} & \textbf{Multi.} \\
    \hline
    \multicolumn{6}{c}{All layers} \\
    \textbf{PTB} & $0.77\ {\pm0.11}$ & $0.79\ {\pm0.11}$ & $0.78\ {\pm0.11}$ & $0.77\ {\pm0.11}$ & $0.73\ {\pm0.20}$ \\
    \textbf{CTB} & $0.77\ {\pm0.11}$ & $0.77\ {\pm0.11}$ & $0.76\ {\pm0.11}$ & $0.74\ {\pm0.11}$ & $0.70\ {\pm0.20}$ \\
    \textbf{KTB} & $0.82\ {\pm0.12}$ & $0.75\ {\pm0.12}$ & $0.79\ {\pm0.12}$ & $0.73\ {\pm0.12}$ & $0.69\ {\pm0.29}$ \\
    \hline
    \multicolumn{6}{c}{Layers longer than 40} \\
    \textbf{PTB} & $0.78\ {\pm0.04}$ & $0.80\ {\pm0.04}$ & $0.77\ {\pm0.04}$ & $0.74\ {\pm0.04}$ & $0.69\ {\pm0.08}$ \\
    \textbf{CTB} & $0.79\ {\pm0.04}$ & $0.80\ {\pm0.04}$ & $0.78\ {\pm0.04}$ & $0.77\ {\pm0.04}$ & $0.76\ {\pm0.07}$ \\
    \textbf{KTB} & $0.90\ {\pm0.04}$ & $0.80\ {\pm0.05}$ & $0.86\ {\pm0.04}$ & $0.77\ {\pm0.06}$ & $0.84\ {\pm0.07}$ \\
    \hline
    \end{tabular}
    }
    \end{center}
    \caption{Mean and standard deviation of compression ratios of Figures \ref{fig:level_ratio_full} \& \ref{fig:level_ratio_multi_full}.
    Longer layers have converged deviations. The last column came from the multi-branching treebanks without a binarizing factor. \label{tab:comp_ratios}}
\end{table*}

Figure~\ref{fig:trees} presents examples for tree binarization and the worst case of $O(n^2)$ complexity.
Figure~\ref{fig:all_complexities} shows the overall linear data complexities in the three languages.
Figures~\ref{fig:level_ratio_full} \& \ref{fig:level_ratio_multi_full} and Table~\ref{tab:comp_ratios} indicate 
that, given a language and a factor, the compression ratio is stable and seldom affected by the sentence length.

The regressions for PTB and CTB show weak $O(n^2)$ tendencies; the quadratic coefficients can be either positive or negative.
Meanwhile, KTB falls into the worst case, as shown in Figure \ref{fig:all_complexities}.
This is because KTB trees tend to have a flat structure on the right side of parses, as illustrated in Figure \ref{fig:zhjp}.
Relaying nodes in the flat structure never combine until the final layer, creating strong $O(n^2)$ tendencies.
As a result, all KTB datasets fall into the worst case, especially when binarized with the CNF-\textbf{left} factor.

A preprocess that groups the flat structure into the \textit{sub} category can prevent considerable quadratic impacts on all datasets.
All $O(n^2)$ tendencies are largely weakened across three corpora, and all linear coefficients drop significantly, as illustrated on the right of Figure \ref{fig:all_complexities}.
The preprocess cannot eradicate the worst case in KTB.
However, all linear coefficients' magnitudes are at least hundreds of times larger than those of the quadratic terms.
In our sub-quadratic case, 200 words lead to approximately 1.5K nodes.
Meanwhile, a sentence with $n$ words has a triangular chart with $\frac{n(n+1)}{2}$ nodes, whose quadratic coefficient is 0.5.
In this case, 200 words lead to approximately 20K nodes.
% This implies the existence of the quadratic upper bound.

\subsection{Experiment Setting} \label{sec:exp_set}

The treebanks PTB and CTB have been widely used for experiments.
For PTB, sections 2-21 were used for training, section 22 for development, and section 23 for testing.
For CTB, articles 001-270 and 440-1151 were used for training, 301-325 for development, and 271-300 for testing.
There is no widely accepted data split for the KTB corpus, except for some probabilistic divisions,
because KTB contains mixed data from sources such as newswires, book digests, and Wikipedia.
We randomly reserved 2,075 samples for development, 1,863 samples for testing, and the remaining 3.3 million as training samples.
Few sentences in the training sets were longer than 100 words
(3 of 40K in PTB; 96 of 17K in CTB; 55 of 33K in KTB).
Frozen English (wiki.en.bin), Chinese (cc.zh.300.bin), and Japanese (cc.ja.300.bin) embeddings 
were used for PTB, CTB, and KTB, respectively\footnote{\url{https://fasttext.cc/}}.
We fed fastText with the PTB text to train \texttt{cbow} instead of \texttt{skipgram} embeddings for \textbf{B}/E with their default settings for 50 epochs.

The batch size was 80, and sentences longer than 100 words were excluded for the triangular data
to avoid out-of-memory (OOM) errors on a single GeForce GTX 1080 Ti with 11 GB.
% We used a TITAN RTX GPU with 24 GB memory to tune our models with XLNet, but tested speeds on 1080 Ti.
We froze XLNet to train our model and then tuned XLNet from the 5-th epoch.
We doubled the batch size at the inference phase to 160.

We used the Adam optimizer with a default learning rate of $10^{-3}$,
while we opted for the XLNet's Adam hyperparameters when tuning the pre-trained XLNet
(e.g., their learning rate was $10^{-5}$).
We adopted a warm-up period for one epoch and a linear decrease
after the 15-th decrease since the last best evaluation.
The recurrent dropout rate was 0.2; other dropout probabilities for FFNNs were set to 0.4.
For model selection, the training process terminated
when the development set did not improve
above the highest score after 100 consecutive evaluations.
The Evalb program\footnote{\url{https://nlp.cs.nyu.edu/evalb/}} was used for F1 scoring.

\begin{table}[t]
    \begin{center}
        \scalebox{.9}{
        \begin{tabular}{l c c r c r}
        \hline
        \textbf{Input} && \multicolumn{2}{c}{\textbf{Development}} & \multicolumn{2}{c}{\textbf{Test}} \\
        \textbf{Comp.} & \textbf{M.} & \textbf{F1} & $P-R$ & \textbf{F1} & $P-R$ \\
        \hline
        \textbf{Frozen}  & \textbf{B} & 92.50 & 0.00 & 92.54 & $+$0.56\\
        \textbf{fastText} & \textbf{M} & 92.10 & $-$0.35 & 92.10 & $-$0.03\\
        \hline
        \textbf{Tuned}  & \textbf{B} & 95.64 & $-$0.05 & 92.72 & $+$0.19 \\
        \textbf{XLNet}  & \textbf{M} & 95.34 & $-$0.15 & 92.44 & $+$0.30 \\
        \hline
        \end{tabular}
        }
    \end{center}
    \caption{F1 scores and differences in precision and recall ($P-R$) on the PTB development and test sets.\label{tab:dev_score}}
\end{table}

We demonstrated score profiles for our main models in Table \ref{tab:dev_score}.
The discrepancy in F1 scores and difference between precision and recall are relatively small on the PTB development and test sets.

\subsection{Variants of Binary Compose} \label{sec:bi}

\begin{algorithm}[t]
    \SetKwFunction{Binary}{BINARY}
    \SetKwFunction{FFNNitp}{FFNN$_{binary}$}
    \SetAlgoLined
    \DontPrintSemicolon
    % \SetNoFillComment
    % \LinesNotNumbered
    \SetInd{0.0em}{1.5em}
    \SetKwProg{Fn}{Function}{:}{}
    \Fn{\Binary{$o_L, o_R,\ x_L, x_R; \text{Var}$}}{
        $x \gets o_L \cdot x_L + (1-o_R) \cdot x_R$\;
        \uIf(\tcp*[f]{relay}){$o_L + (1-o_R) = 1$}{
            \KwRet $x$\;
        % }\uElseIf{Combine is Add:Mul(b)}{
        %     $\hat{x}_{[<b]}^{mul} \gets x_{L[<b]} \odot x_{R[<b]}$
        %     \tcp*{mul region}
        %     $\hat{x}_{[\geq b]}^{add} \gets x_{L[\geq b]} + x_{R[\geq b]}$
        %     \tcp*{add region}
        %     $\hat{x} \gets \hat{x}_{[<b]}^{mul} \oplus \hat{x}_{[\geq b]}^{add}$
        %     \tcp*{concat regions}
        %     $x \gets \hat{x} \odot \sigma$ \FFNN$_{post}${($\hat{x}$)}
        %     \tcp*{post gate}
        }\uElseIf(\tcp*[f]{\textit{ADD}}){Var is \textit{ADD}}{
            \KwRet $x$\;
        }
        \uElse{
            \uIf(\tcp*[f]{\underline{N}o input}){Var is \underline{N}S or \underline{N}V}{
                $\lambda \gets \sigma$ \FFNNitp{$\emptyset$}\;
            }\uElseIf(\tcp*[f]{\underline{C}oncat$\dots$}){Var is \textit{\underline{C}S} or \textit{\underline{C}V}}{
                $\lambda \gets \sigma$ \FFNNitp{$x_L \oplus x_R$}\;
            }\uElseIf(\tcp*[f]{\underline{B}iaffine}){Var is \textit{\underline{B}V}}{
                $\lambda \gets \sigma$ \FFNNitp{$x_L, x_R$}\;
            }
            \uIf(\tcp*[f]{\underline{V}ector $\lambda$}){Var is \textit{N\underline{V}}, \textit{C\underline{V}}, or \textit{B\underline{V}}}{
                $x \gets \lambda \odot x_L + (1-\lambda) \odot x_R$\;
            }
            \uElseIf(\tcp*[f]{\underline{S}calar $\lambda$}){Var is \textit{N\underline{S}} or \textit{C\underline{S}}}{
                $x \gets \lambda \cdot x_L + (1-\lambda) \cdot x_R$\;
            }
            \KwRet $x$
            \tcp*{\textit{NS NV CS CV BV}}
        }
    }
\caption{Binary Compose Variants\label{alg:bin_var}}
\end{algorithm}

If we choose the relay instruction in line 12 of Algorithm \ref{alg:enc_bin}, 
additive vector compositionality is retained~\cite{DBLP:conf/nips/MikolovSCCD13}
as the na\"{\i}ve \textbf{ADD} variant in lines 5--6 of Algorithm \ref{alg:bin_var}.
The model can infer a full tensor tree;
however, \textbf{ADD} causes the vector magnitude to increase with the tree height cumulatively.
This is unwanted in the recurrent or recursive neural network.

Therefore, we examined a learnable \texttt{FFNN}$_{multi}$ with Sigmoid activation 
to perform gate-style interpolation in five variants \textbf{NS}, \textbf{NV},
\textbf{CS}, \textbf{CV}, and \textbf{BV} as described in lines 8--17.
When a variant takes \underline{n}o input and 
produces a \underline{s}calar interpolation parameter $\lambda$, we consider this case \textbf{NS}.
(``$\varnothing$'' is a placeholder for no input.)
Meanwhile, \textbf{CV} indicates \underline{c}oncatenated input and \underline{v}ectorized interpolation.
\textbf{BV} is a variant that involves a \underline{b}iaffine tensor operation.
\textbf{CV} is our default \texttt{BINARY} variant;
the experiments for these variants are presented in Table \ref{tab:bin}.

\begin{table}[t]
    \begin{center}
        \scalebox{.9}{
        \begin{tabular}{c p{4.7cm} c}
        \hline
        \textbf{Var} & \centering \textbf{Specification} & \textbf{F1} \\
        \hline
        \textbf{BV} & Biaffine inputs for vector $\lambda$. & \underline{92.53} \\
        \textbf{CV} & $x_L\oplus x_R$ as input for vector $\lambda$. & \textbf{92.54} \\
        \textbf{CS} & $x_L\oplus x_R$ as input for scalar $\lambda$. & 91.83 \\
        \textbf{NV} & No input; bias vector $\lambda$. & 92.36 \\
        \textbf{NS} & No input; bias scalar $\lambda$. & 91.95 \\
        \textbf{ADD} & \centering $x_L + x_R$ & 91.86 \\
        % \textbf{AVG} & \centering $x = \frac{x_L + x_R}{2}$ & 91.48 \\
        % \textbf{SOR} & \centering $x = \frac{x_L + x_R}{1.5+0.5\cdot\cos(4\langle x_L, x_R\rangle + \pi)}$ & 91.96 \\
        % \textbf{HOR} & For the same sign, pointwisely take either $x_L$ or $x_R$ by the maximum absolute value, otherwise addition. & 91.86 \\
        \hline
        \end{tabular}
        }
    \end{center}
    \caption{Compositionality of the \texttt{BINARY} function.\label{tab:bin}}
\end{table}

In terms of the F1 score, the most competitive variants of \textbf{CV} 
are \textbf{BV} and \textbf{NV}, suggesting that fine interpolation can effectively facilitate vector compositionality. 
The similarity in results of \textbf{CS}, \textbf{NS}, and \textbf{ADD} validate this suggestion.
This indicates that vector compositionality is not as trivial as an additive function at the scalar level,
and a matrix operation is sufficient.
\textbf{BV} is the costliest variant with a tensor operation that runs very slowly (30 sents/sec).

% We examined the combinatory functions; the results are presented in the bottom rows of Table \ref{tab:abl}. % of Table \ref{tab:abl}.
% \textbf{CV} involves a matrix operation, whereas \textbf{CS}, \textbf{NV}, and \textbf{NS} are truncated variants with vector operations.

\subsection{Recovering Symbolic Tree} \label{sec:symbolic}

\begin{algorithm}[t]
    \SetKwFunction{Symbolic}{REC}
    \SetKwFunction{Tree}{TREE}
    \SetAlgoLined
    \DontPrintSemicolon
    \SetInd{0.0em}{1.2em}
    \SetKwProg{Sn}{Function}{:}{}
    \Sn{\Symbolic{$x_{0:n},\ t_{0:n},\ l^{k}_{0:n_k},\ o^{k}_{0:n_k}\ or \ c^{k}_{0:n_k+1}$}}{
        \For{$i\gets0$ \KwTo $n-1$}{
            $tree_i \leftarrow$ \Tree{$t_i, x_i$}\;
        }
        \For{$j\gets 0$ \KwTo $k-2$}{
            \eIf(\tcp*[f]{\texttt{BINARY}}){binary parsing}{
                \For{$i\gets 0$ \KwTo $n_j-1$}{
                    \uIf{$o_i^{j} + (1-o_{i+1}^{j}) = 2$}{
                        Combine $tree_{i:i+2}$ under $l_{\mathit{parent\_of(i,i+1)}}^{j+1}$\;
                    }
                }

            }(\tcp*[f]{\texttt{MULTI-BRANCHING}}){
                \ForEach{$chk\ \mathit{in}\ c^j_{0:n_j+1}$}{
                    Combine $tree_{chk}$ under $l_{\mathit{parent\_of(chk)}}^{j+1}$\;
                }
            }
        }
        Expand unary and flatten \textit{sub} labels for $tree_0$\;
        \KwRet{$tree_0$}
    }
    \caption{Recovering a Symbolic Tree\label{alg:dec}}
\end{algorithm}

To obtain the final tree representation, we initialized the working place with leaves of words and predicted POS tags.
Two symbolic rules were used to modify the labels and construct sub-trees, as described in Algorithm \ref{alg:dec}.
1) The collapsed unary branches were expanded to their original structure by splitting at the plus marks 
(e.g., \texttt{SBAR+S} into \texttt{SBAR} and \texttt{S}).
2) `the label is a \textit{sub}' excluded the repeated labels and relayed sub-trees.
These rules enabled a single $tree_0$ as the final output.

\subsection{Chinese and Japanese Headedness} \label{sec:zhjp}

\begin{figure}[t]
    \centering
    \includegraphics[width=\linewidth]{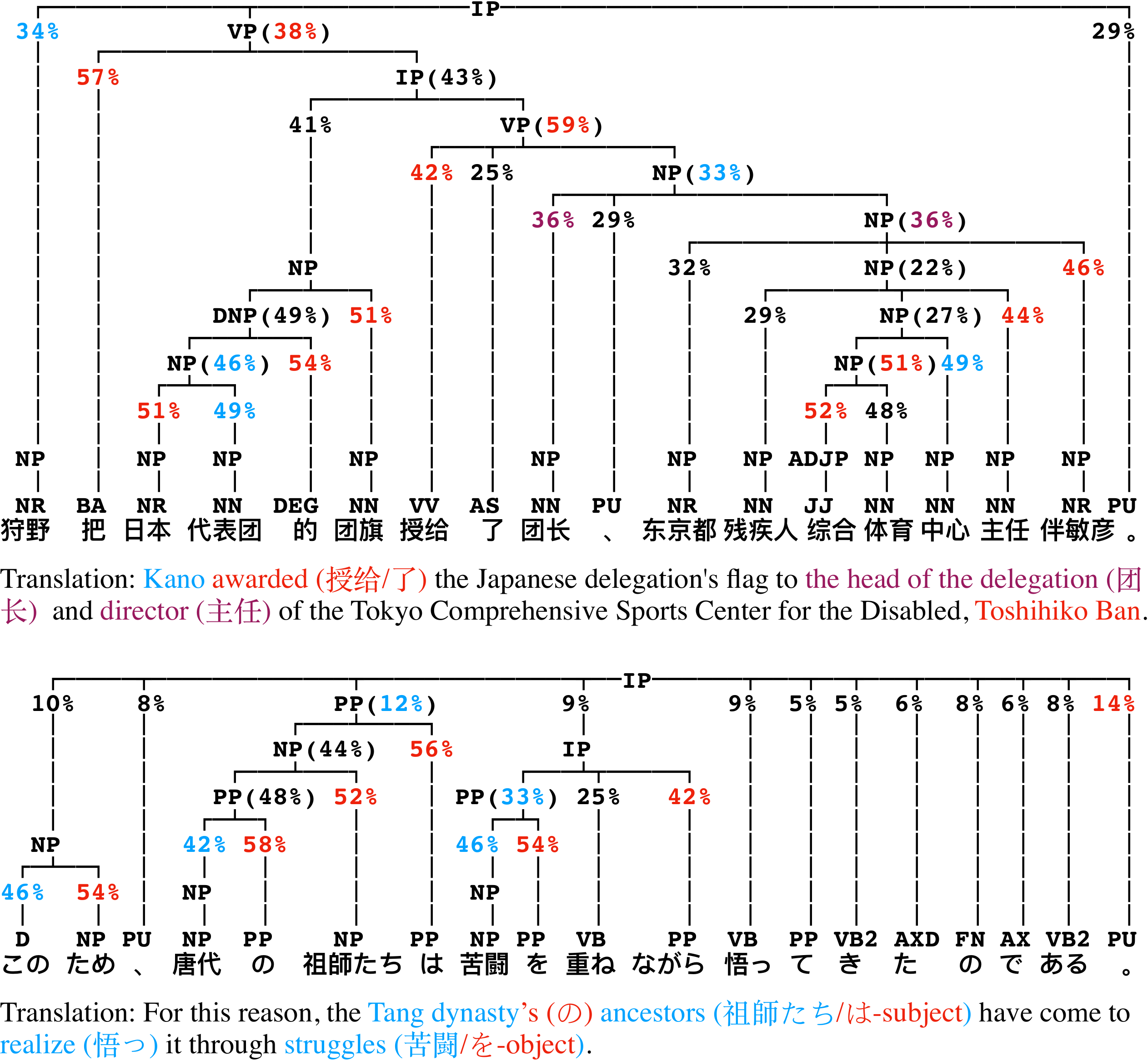}
    \caption{
        Chinese (top) and Japanese (bottom) parses from the multi-branching model. \label{fig:zhjp}}
\end{figure}

Figure \ref{fig:zhjp} presents two non-English parses from the multi-branching model.
Both the Chinese and Japanese languages possess functional markers that receive high attention (percentage and words in red),
such as the second character tagged with \texttt{BA} in Chinese, and Japanese case markers tagged with \texttt{PP}.
Interestingly, the Chinese verb (i.e., the one meaning ``awarded'') received the highest attention,
whereas Japanese verbs (i.e., two sub-words tagged with \texttt{VB}) did not.
We supposed the reason behind this is that Japanese sentences drop the \texttt{VB}s and other heads more often than Chinese.
The coordinated \texttt{NP}s in the Chinese parse (i.e., two words meaning ``head'' and ``director'') received equal attention weights.

Moreover, two trees show their branching tendencies: Chinese is \textbf{midin}-alike; Japanese is a \textbf{left}-branching language, 
and KTB has a large flat structure on the right.

\end{document}